\documentclass{article} 
\usepackage{iclr2025_conference,times}

\usepackage{amsmath,amsfonts,bm}

\def\eqref#1{equation~\ref{#1}}
\def\Eqref#1{Equation~\ref{#1}}

\def\1{\bm{1}}

\DeclareMathAlphabet{\mathsfit}{\encodingdefault}{\sfdefault}{m}{sl}
\SetMathAlphabet{\mathsfit}{bold}{\encodingdefault}{\sfdefault}{bx}{n}

\usepackage{hyperref}
\usepackage{url}
\usepackage{booktabs}       
\usepackage{amsfonts}       
\usepackage{nicefrac}       
\usepackage{microtype}      
\usepackage{xcolor}         
\usepackage{amsmath}
\usepackage{latexsym}
\usepackage{pifont}%
\usepackage{amsbsy}
\usepackage{amssymb}
\def\eqref#1{(\ref{#1})}
\usepackage{graphicx}
\usepackage{subcaption}
\usepackage{enumerate}
\usepackage{enumitem}
\usepackage{mathtools}
\usepackage{color}
\usepackage{float}
\usepackage{makecell}
\usepackage{arydshln}
\usepackage{wrapfig}
\usepackage{arydshln} 

\hypersetup{
    colorlinks=true,        
    citecolor=teal,         
    linkcolor=violet,       
    urlcolor=blue           
}

\usepackage{multirow}
\newtheorem{theorem}{Theorem}

\newtheorem{proposition}[theorem]{Proposition}

\newcommand{\cmark}{\ding{51}}%
\newcommand{\xmark}{\ding{55}}%

\title{MallowsPO: Fine-Tune Your LLM with Preference Dispersions}

\author{Haoxian Chen$^1$\thanks{Equal Contribution, $\alpha$-$\beta$; Email: \{\texttt{hc3136,hz2684,khl2114,yao,wt2319}\}@columbia.edu}\text{ }\hspace{0.8mm}, Hanyang Zhao$^1$$^*$, Henry Lam$^1$, David D. Yao$^1$, Wenpin Tang$^1$\\
$^1$Columbia University, Department of IEOR\\
}

\iclrfinalcopy 
\begin{document}

\maketitle

\begin{abstract}
Direct Preference Optimization (DPO) has recently emerged as a popular approach to improve reinforcement learning from human feedback (RLHF), leading to better techniques to fine-tune large language models (LLM). A weakness of DPO, however, lies in its lack of capability to characterize the diversity of human preferences. 
Inspired by Mallows' theory of preference ranking, we develop in this paper a new approach, the {\it MallowsPO}. 
A distinct feature of this approach is a {\it dispersion index}, which reflects the dispersion of human preference to prompts. We show that existing DPO models can be reduced to special cases of this dispersion index, thus unified with MallowsPO. More importantly, we demonstrate empirically how to use this dispersion index to enhance the performance of DPO in a broad array of benchmark tasks, from synthetic bandit selection to controllable generation and dialogues, while maintaining great generalization capabilities.
MallowsPO is also compatible with other SOTA offline preference optimization methods, boosting nearly 2\% extra LC win rate when used as a plugin for fine-tuning Llama3-Instruct.
\end{abstract}

\section{Introduction}

Reinforcement Learning from Human Feedback (RLHF) \citep{RLHF2022,LSHF2020,FTLM2020} has made significant 
contributions to the success of 
Large Language Models (LLMs) such as \texttt{ChatGPT} and \texttt{GPT4} \citep{GPT4}. 
Recently, Direct Preference Optimization (DPO) \citep{DPO}
is proposed as a way to bypass RL, thus leading to faster training and better resource efficiency. 
More importantly, DPO achieves comparable or superior performance against RLHF in downstream tasks such as fine-tuning LLMs in Llama3 \citep{Llama3}, Zephyr \citep{Zephyr}, BTLM-DPO \citep{Tulu2}, etc. DPO's success has attracted much research attention, leading to variants 
beyond pairwise ranking such as KTO \citep{KTO}, 
unified perspectives on loss parameterization such as IPO \citep{IPO}, GPO \citep{GPO}, and reference-free alternatives such as CPO \citep{CPO}, ORPO \citep{ORPO}, SimPO \citep{SimPO}, etc. See \citep{winata2024preference} for a comprehensive survey.

\begin{figure}[!htbp]
  \centering
  \caption{\small{Prompts with low/high neg-log dispersion estimate values from Anthropic HH dataset}.}
    \includegraphics[width=0.79\linewidth]{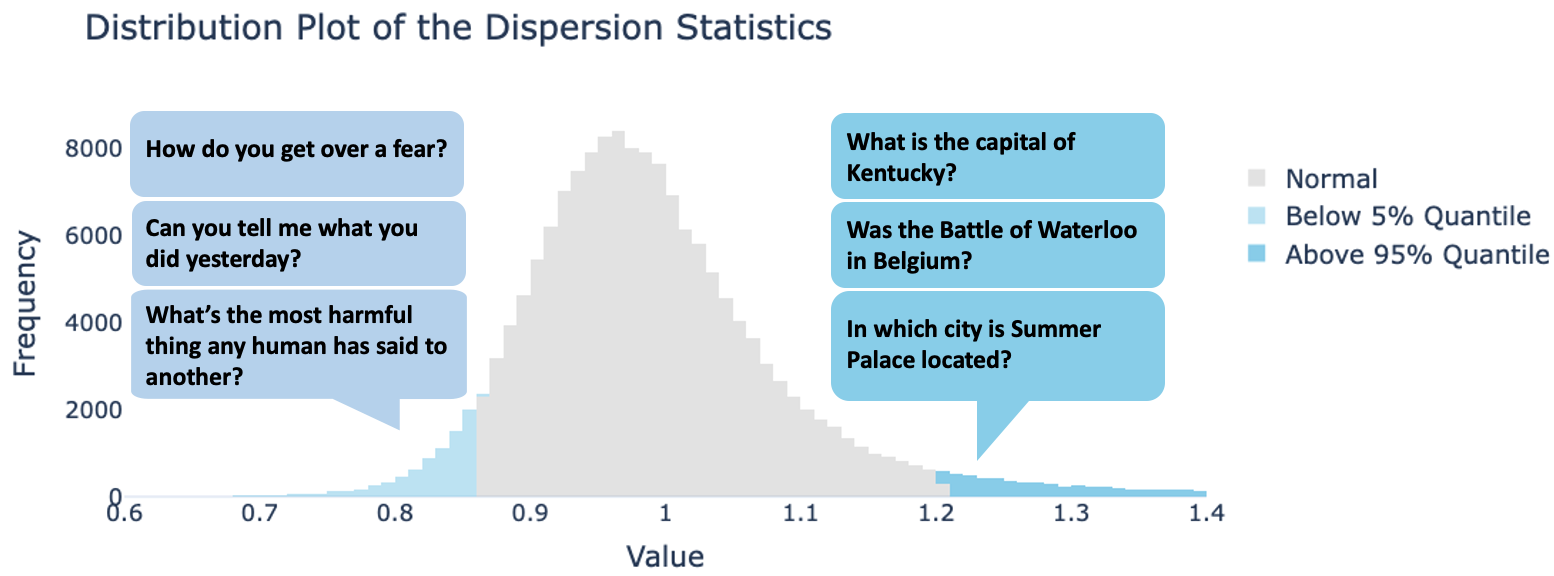}
    
    \vspace{-15pt}
    \label{fig:prompt dispersion with examples}
\end{figure}

Notwithstanding the successes by RLHF and DPO, both are limited by the assumption that preference follows the Bradley-Terry (BT) model \citep{BT}. In particular, this does not account for varying degrees of agreement in response to different prompts. For instance, people are more likely to agree on ``$1+1=$? // $2$.'' as opposed to ``What is the best city to live in the U.S.? // New York.''  In language models, this concerns the {\em dispersion} of next-token prediction, which is reminiscent to {\em personalization} in recommendation systems \citep{Chan22, fu2022there}. See Figure \ref{fig:prompt dispersion with examples} for more illustrative examples of prompts in Anthropic HH \citep{AnthropicHH} dataset.

The purpose of this paper is to formalize the idea of prompt dispersion in the design of DPO.  
We adapt Mallows' preference ranking theory \citep{diaconis1988group, mallows}, 
a family of ranking models that provide a natural carrier for prompt dispersion, and
propose the following decomposition/factorization of the (latent) reward function:
\begin{equation*}
\mbox{reward}(\mbox{prompt, completion}) = 
\mbox{dispersion}(\mbox{prompt}) \times \mbox{scaled reward}(\mbox{completion} \mid \mbox{prompt}),
\end{equation*}
where ``prompt'' and ``completion'' correspond, respectively, to question and answer. 
This decomposition allows us to specify
the diverse level of prompt dispersions hidden in the DPO,
which is translated into a prompt-dependent factor -- the {\em dispersion index} -- in the preference likelihood. 
The scaled reward is given by the relative rank of the (possible) completions, which further enhances model interpretability. 
We then leverage the change of variables technique to propose two models, MallowsPO-$\theta$ and MallowsPO-$\phi$, 
by two choices of the discrepancy function in the Mallows model 
which we will elaborate in Section \ref{sec:mallows_intro}.

\vspace{-5pt}

The main contributions of this paper are three-fold.

(1) We formalize the idea of prompt dispersion in DPO, and develop the MallowsPO approach to implement this idea, so as to improve and generalize DPO. 
To the best of our knowledge, this is the first work in 
preference optimization that considers a mathematically well-grounded preference ranking model (Mallows) beyond the BT model.

(2) We propose approximations to the dispersion index, a critical component of the Mallows model, so as to facilitate computation and validate the approximation through a synthetic example. This allows Mallows model to serve as a foundation for developing versatile preference optimization objectives, including MallowsPO-$\theta$ (a generalization of DPO) and MallowsPO-$\phi$. We also provide various analytical results for MallowsPOs, uncovering various new insights on existing DPO models, and a generalized $\Psi$PO model that unifies all DPO models (including MallowsPO).

(4) We conduct extensive experiments, from synthetic bandits, controllable generation, fine-tuning Pythia 2.8B on off-policy Anthropic HH dataset, to fine-tuning Llama3-8B-Instruct on a on-policy UltraFeedback prompts based dataset. 
Notably, we perform an exclusive hyperparameter search for a fair comparison, and repeat for different random seeds to justify the significance of the improvement. The results show clear advantages of MallowsPO over (BT-)DPO, highlighting the potential of considering preference/prompt dispersion.

\textbf{Related Works}.
Existing work on personalization in dialogue generation such as \citet{fu2022there} and \citet{li2016deep} has also paid attention to the diversity of human preferences (``there are a thousand Hamlets in a thousand people's eyes'');
\cite{Nashing_Learning} proposes a Nash game model to incorporate the diversity. 
There are also other DPO variants:
$f$-DPO \citep{f-DPO} considers general $f$-divergence 
in DPO; 
ODPO \citep{ODPO} adds a margin to account for the preference significance.
Recent works propose to learn online preferences \citep{calandriello2024human,tajwar2024preference}, 
or learn from AI feedbacks \citep{bai2022constitutional,chen2024self,lee2023rlaif}.
For classical RLHF, studies to improve the design and capabilities of RLHF include \citet{Alpacafarm,kirk2023understanding, Secrets_of_RLHF_II,zhai2023investigating,zhao2023slic,Secrets_of_RLHF_I}, whose ideas can also benefit DPO. 
See \cite{winata2024preference} for a survey on learning from preferences.

The remainder of the paper is organized as follows.
Background materials on RLHF and DPO are highlighted in Section \ref{sc2}. 
Section \ref{sc3} focuses on the development of MallowsPO,
followed by more analytical results and various perspectives in Section \ref{sc4}.
Experimental results are detailed in Section \ref{sc5}, 
and concluding remarks in Section \ref{sc6}.

\section{Preliminaries}
\label{sc2}

Both RLHF  and DPO 
start with fine-tuning a pre-trained LLM by supervised learning on
high-quality data for some downstream tasks of interest,
to acquire a model $\pi^{\mathrm{SFT}}$.
This step is referred to as the supervised fine-tuning (SFT) phase. 
For instance, for training InstructGPT \citep{RLHF2022},
GPT-3 \citep{GPT3} is first fine-tuned on the given input prompt distribution.

$\diamondsuit$ \textbf{RLHF} \citep{RLHF2022,LSHF2020,FTLM2020}. 
On top of $\pi^{\mathrm{SFT}}$, RLHF is proposed to serve as the next step to conduct further fine-tuning to generate high-quality outputs as judged by humans.
Given a generative model $\pi$, it is prompted with prompts $x$ to produce pairs of answers (or, ``completions''),
$\left\{y_1, y_2\right\} \sim \pi(y \mid x)$,
which are then presented to human labelers 
who express preferences for one completion over the other.
Denote by $y_w \succ y_l \mid x$,
meaning that $y_w \in \left\{y_1, y_2\right\}$ is preferred over $y_l \in \left\{y_1, y_2\right\}$. 
The preferences are assumed to be generated by some latent reward model $r^*(x, y)$.
Based on the collected preference data $\{x^{(i)}, y_w^{(i)}, y_l^{(i})\}_{i=1}^N$, RLHF consists of first learning a reward model $r(x,y)$, 
followed by learning a policy $\pi_r (y \mid x)$ 
in which the prompt $x$ is the state, and the completion $y$ is the action.

(a) {\bf Reward model}.
To capture the underlying human preferences, 
RLHF assumes the Bradley-Terry model \citep{BT} that stipulates the pairwise preference distribution:
\begin{equation}
\label{eq:BT}
p^*\left(y_1 \succ y_2 \mid x\right):=
\sigma\left(r^*\left(x, y_1\right) - r^*\left(x, y_2\right) \right),
\end{equation}
where $\sigma(s):=\frac{1}{1+ e^{-s}}$.
Given access to a static dataset of comparisons
$\mathcal{D}=\{x^{(i)}, y_w^{(i)}, y_l^{(i)}\}_{i = 1, \ldots, N}$,
RLHF seeks to approximate the latent reward $r^*(x,y)$ by a family of functions $\{r_\psi(x,y)\}_\psi$,
and estimate the parameters by minimizing the (negative) log-likelihood loss:
$\min_\psi \mathcal{L}\left(r_\psi, \mathcal{D}\right):=-\mathbb{E}_{\left(x, y_w, y_l\right) \sim \mathcal{D}}\left[\log \sigma\left(r_\psi\left(x, y_w\right)-r_\psi\left(x, y_l\right)\right)\right]$.
Denote by $r_{\psi_*}(x,y)$ the solution to the problem.

(b) {\bf RL}. The learned reward function $r_{\psi_*}(x,y)$ is then used to provide feedback to the language model.
More precisely, the following  KL-regularized RL problem is considered:
\begin{equation}
\label{RLHF objective}
\begin{aligned}
& \max_{\pi} \mathbb{E}_{x \sim \mathcal{D}}\left[\mathbb{E}_{y \sim \pi (y \mid x)}\left[r_{\psi_*}(x, y)\right]-\beta \mathrm{KL}\left(\pi (\cdot\mid x) \| \pi_{\mathrm{ref}}(\cdot \mid x)\right)\right],
\end{aligned}
\vspace{-5 pt}
\end{equation}
where $\beta > 0$ is a hyperparameter controlling the deviation from the reference policy
$\pi_{\mathrm{ref}} = \pi^{\mathrm{SFT}}$.
In view of \eqref{RLHF objective}, RLHF uses the reward function
$r(x, y)=r_\psi(x, y)-\beta\left(\log \pi (y \mid x)-\log \pi_{\text {ref }}(y \mid x)\right)$,
and solves the RL problem by proximal policy optimization (PPO) \citep{PPO}.

$\diamondsuit$ \textbf{DPO} \citep{DPO}.
One disadvantage of RLHF is that the RL step often requires substantial computational effort (e.g., to carry out PPO). 
The idea of DPO is to combine the two steps (a)--(b) in RLHF into a single one,
bypassing the computation in the RL step.

The key idea is that given a reward function $r(x,y)$,
the problem in \eqref{RLHF objective}
has a closed-form solution:
$\pi_r(y \mid x)=\frac{1}{Z(x)} \pi_{\text {ref }}(y \mid x) \exp \left(r(x, y)/\beta\right)$,
where $Z(x)=\sum_y \pi_{\text {ref }}(y \mid x) \exp \left( r(x, y)/\beta\right)$.
Rewrite the above as:
$r(x, y)=\beta \log \frac{\pi_r(y \mid x)}{\pi_{\text {ref }}(y \mid x)}+\beta \log Z(x)$.
Through this change of variables, the latent reward $r^*(x,y)$ can be expressed in terms of the optimal
 policy $\pi^*(y \mid x)$,
the reference policy $\pi_{\text {ref }}(y \mid x)$
and a constant $Z^*(x)$.
Substituting this $r^*$ expression into \eqref{eq:BT} yields:
\begin{equation}
\label{eq:ppi}
p^*\left(y_1 \succ y_2 \mid x\right)=\sigma\left(\beta \log \frac{\pi^*\left(y_1 \mid x\right)}{\pi_{\text {ref }}\left(y_1 \mid x\right)}-\beta \log \frac{\pi^*\left(y_2 \mid x\right)}{\pi_{\text {ref }}\left(y_2 \mid x\right)}\right),
\end{equation}
where $Z^*(x)$ cancels out. 
The expression in \eqref{eq:ppi} motivates the DPO objective:
\begin{equation}
\label{DPO objective}
\min_{\pi} \mathcal{L}_{\mathrm{DPO}}\left(\pi ; \pi_{\mathrm{ref}}\right)
:=-\mathbb{E}_{\left(x, y_w, y_l\right) \sim \mathcal{D}}\left[\log \sigma\left(\beta \log \frac{\pi \left(y_w \mid x\right)}{\pi_{\text {ref }}\left(y_w \mid x\right)}-\beta \log \frac{\pi \left(y_l \mid x\right)}{\pi_{\text {ref }}\left(y_l \mid x\right)}\right)\right].
\end{equation}

\section{DPO based on Mallows Ranking Models}
\label{sc3}

\subsection{Mallows ranking models}\label{sec:mallows_intro}
The Mallows model is built upon the analysis of rankings,
instead of scores or ratings that play the central role in BT models.
Concretely, for a positive integer $n$ that represents e.g., $n$ possible items, let $\mathfrak{S}_n$ be the set of permutations of $[n]=\{1, \ldots, n\}$ and the space of rankings. 
We consider that the probability of observing a ranking $\mu$ (which represents the preference of $n$ items, e.g., the top ranked item is preferred over the others) follows a exponential family characterized as:
\begin{equation}
\label{preference probability}
\mathbb{P}_{\phi, \mu_0, d}(\mu):= \phi^{d\left(\mu, \mu_0\right)}/Z(\phi, d) \quad \text { for } \mu \in \mathfrak{S}_n,
\end{equation}
where $\phi \in (0,1]$ is the dispersion parameter,
$\mu_0$ is the central ranking,
$d(\cdot, \cdot)$ is a discrepancy function that is right invariant: $d(\mu_1, \mu_2)=d\left(\mu_1 \circ \mu_2^{-1}, i d\right)$ for $\mu_1, \mu_2 \in \mathfrak{S}_n$,
and $Z(\phi, d):=\sum_{\mu \in \mathfrak{S}_n} \phi^{d\left(\mu, \mu_0\right)}$
is the normalizing constant. Intuitively, the Mallows model is probabilistic ranking model in which the probability of a rank decreases exponentially with its distance from the central rank. The greater the difference between a rank $\mu$ and the central rank, the lower its probability.

Notice that the dispersion indeed reflects how dispersed the probability distribution defined by \eqref{preference probability} on the space of rankings is: When $\phi \to 0$, it is point pass concentrated on $\mu_0$,
and when $\phi = 1$, it is uniformly distributed.
In an attempt to study ranking models (over $n$ items) with pairwise preferences,
\cite{mallows} further considered two specific cases of the discrepancy function in \eqref{preference probability}:

$\bullet$ Mallows-$\theta$ model: $d(\mu_1, \mu_2)=\sum_{i=1}^n(\mu_1(i)-\mu_2(i))^2$ is the Spearman's rho,

$\bullet$ Mallows-$\phi$ model: $d(\mu_1, \mu_2)=\operatorname{inv}\left(\mu_1 \circ \mu_2^{-1}\right)$ is the Kendall's tau,

where $\operatorname{inv}(\mu):=\#\left\{(i, j) \in[n]^2: i<j\right.$ and $\left.\mu(i)>\mu(j)\right\}$ is the number of inversions of $\mu$. 

\textbf{Example.} 
Consider the central ranking of three items being $\mu_0 = (1,2,3)$, 
with the observation $\mu = (3,2,1)$. 
If the underlying model is Mallows-$\theta$ or Mallows-$\phi$, 
we have
$$
d_\theta(\mu,\mu_0) = (3-1)^2 +(2-2)^2 +(1-2)^2 = 4; \quad d_\phi(\mu,\mu_0) = \operatorname{inv}(3,2,1) = 3.$$ 
See more examples and visual plots  in Appendix \ref{app:Mallows Examples}. Notably, different discrepancy functions are usually not consistent, thus they define different Mallows Ranking Models, see an example in \ref{app: inconsistency between Mallows Ranking Models}.

As a remark, the general form in \eqref{preference probability} was suggested by \citet{diaconis1988group} along with other discrepancy functions (e.g., Cayley, Ulam distances, etc.)
See \cite{critchlow2012metric, diaconis1988group, diaconis1989generalization} for the related group representation approach to ranked, or partially ranked data.
Note that the Mallows models 
can be extended to infinite ranking models with $n = \infty$
(see \cite{meila2010exponential, PT19, tang2019mallows}.)
In the context of language models,
this conforms to a possibly infinite number of completions given a prompt,
and allows interpreting unseen completions.

\subsection{MallowsPO} 
We adapt Mallows ranking models highlighted above to the setting of language models. 
First, denote by $\mu(\cdot \mid x)$ a ranking of completions given the prompt $x$, such that the preference distribution is:
\begin{equation}
\label{muprefer}
p^*\left(y_1\succ y_2\mid x\right) =\mathbb{P}\left(\mu(y_1\mid x)<\mu\left(y_2\mid x\right)\right). 
\end{equation}
Next, for the preference probability in \eqref{preference probability}, 
given an input prompt $x$, we assume it induces a conditional central ranking $\mu_0(\cdot\mid x)$,
and a dispersion index $\phi(x) \in (0,1]$.
As pointed out in \cite{tang2019mallows}, 
finding $\mu_0(\cdot \mid x)$ is computationally hard. 
Fortunately, we discover that in RLHF, this part can be ``cleverly" circumvented. By representing  $r^*(x,y)$ as the (negative) rank $-\mu_0(y \mid x)$,
our goal now becomes:
\begin{equation}
\label{Mallows RLHF objective}
\max _{\pi} \mathbb{E}_{x \sim \mathcal{D}}\left[\mathbb{E}_{y \sim \pi_\theta(y \mid x)}\left[-\mu_0(y \mid x)\right]-\beta \mathrm{KL}\left(\pi(\cdot\mid x) \| \pi_{\mathrm{ref}}(\cdot \mid x)\right)\right],
\end{equation}
Note that a {\it smaller} rank is preferred as per \eqref{muprefer}. Hence, this provides a natural candidate for the scaled reward that enhances model interpretation.
This perspective leads to the discovery of a novel family of preference optimization objectives, each of which corresponds to an instance of Mallows models.
\vspace{-5 pt}
\paragraph{MallowsPO-$\theta$.} For Mallows-$\theta$ model, we show by Proposition \ref{prop:Mallows-theta consistent pairwise prob} that, if we have the paired-comparison probability of ranking item $y_1$ before item $y_2$ as
\begin{align}
    \mathbb{P}\left(\mu(y_1\mid x)<\mu\left(y_2\mid x\right)\right)
&= \sigma(2\log\phi(x)(\mu_0(y_1|x) - \mu_0(y_2|x))) \coloneq g_{\theta,\phi(x)} (s)\label{eq:mallows_theta_prob},
\vspace{-10 pt}
\end{align}
where $\log \phi(x)\in (-\infty, 0)$ and $s:=\mu_0\left(y_2\mid x\right)-\mu_0\left(y_1\mid x\right)$, then under Spearman's distance $d$ and given $\phi(x)$, we can recover (\ref{preference probability}) under the assumptions of consistent pairwise probabilities. 

\begin{proposition}[Probability of rank $\mu$ in Mallows-$\theta$]
\label{prop:Mallows-theta consistent pairwise prob}
$~$
Suppose that $\mathbb{P}\left(\mu(y_1\mid x)<\mu\left(y_2\mid x\right)\right)$ satisfies (\ref{eq:mallows_theta_prob}) with given $\phi$ and central ranking $\mu_0$, then we have $\mathbb{P}(\mu) \propto \phi(x)^{\sum_{i=1}^n (\mu(i)-\mu_0(i))^2}$, i.e. $\mu$ is drawn from Mallows-$\theta$ (with Spearman's rho distance $d(\mu_1, \mu_2)=\sum_{i=1}^n(\mu_1(i)-\mu_2(i))^2$).
\end{proposition}

The (inconsistent) pairwise probability in (\ref{eq:mallows_theta_prob}), which yields Mallows-$\theta$ model, leads to our following optimization objective.

\begin{theorem}[MallowsPO-$\theta$] 
\label{thm:mallowspo_theta}
$~$
Given a prompt $x$, its associated dispersion $\phi(x)$ and the underlying central rank $\mu_0(\cdot|x)$, suppose for any $y_1, y_2$ the preference probability satisfies
$p^*\left(y_1\succ y_2\mid x\right)$ as (\ref{eq:mallows_theta_prob}), and thus the preference distribution of $y$ is Mallows-$\theta$. Then, the optimal RLHF policy $\pi_{\mu_0}(\cdot \mid x)$ for (\ref{Mallows RLHF objective}) satisfies $p^*\left(y_1\succ y_2\mid x\right) = \sigma \left(-2\log\phi(x)\left(\beta \log \frac{\pi\left(y_w \mid x\right)}{\pi_{\text {ref }}\left(y_w \mid x\right)}-\beta \log \frac{\pi\left(y_l \mid x\right)}{\pi_{\text {ref }}\left(y_l \mid x\right)}\right)\right)$. 
Therefore, the MallowsPO-$\theta$ policy optimization objective is:
\vspace{-5 pt}
\begin{equation}
\label{Mallows DPO-theta objective}
\min_{\pi} -\mathbb{E}_{\left(x, y_w, y_l\right) \sim \mathcal{D}}\log \sigma (\underbrace{-2\log\phi(x)}_{\text{neg-log dispersion of }x}\left(\beta \log \frac{\pi\left(y_w \mid x\right)}{\pi_{\text {ref }}\left(y_w \mid x\right)}-\beta \log \frac{\pi\left(y_l \mid x\right)}{\pi_{\text {ref }}\left(y_l \mid x\right)}\right)).
\end{equation}
\vspace{-15 pt}
\end{theorem}

The proof of Proposition \ref{prop:Mallows-theta consistent pairwise prob} and Theorem \ref{thm:mallowspo_theta} are given in Appendix \ref{sec:proofs}.

\begin{figure}[!htbp]
  \centering
    \begin{minipage}[t]{0.55\linewidth}
    \centering
    \includegraphics[width=0.8\linewidth]{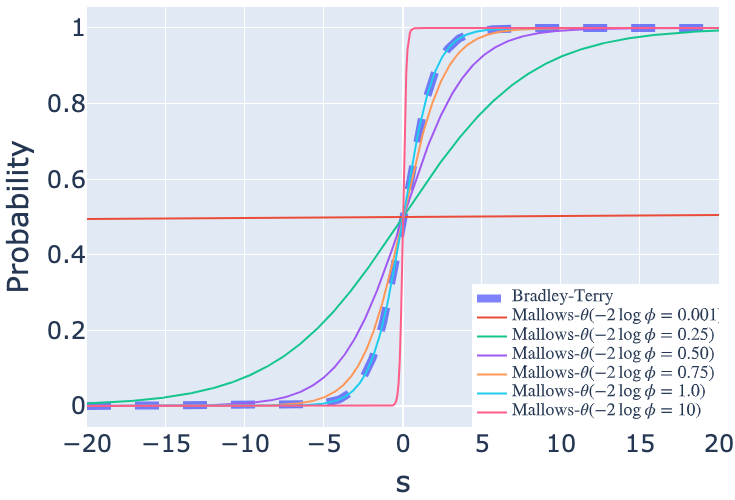}
    \caption{\small{Distribution plot}.}
    \label{fig:dist_function}
  \end{minipage}
  \hfill
  \begin{minipage}[t]{0.44\linewidth}
    \centering
    \includegraphics[width=0.85\linewidth]{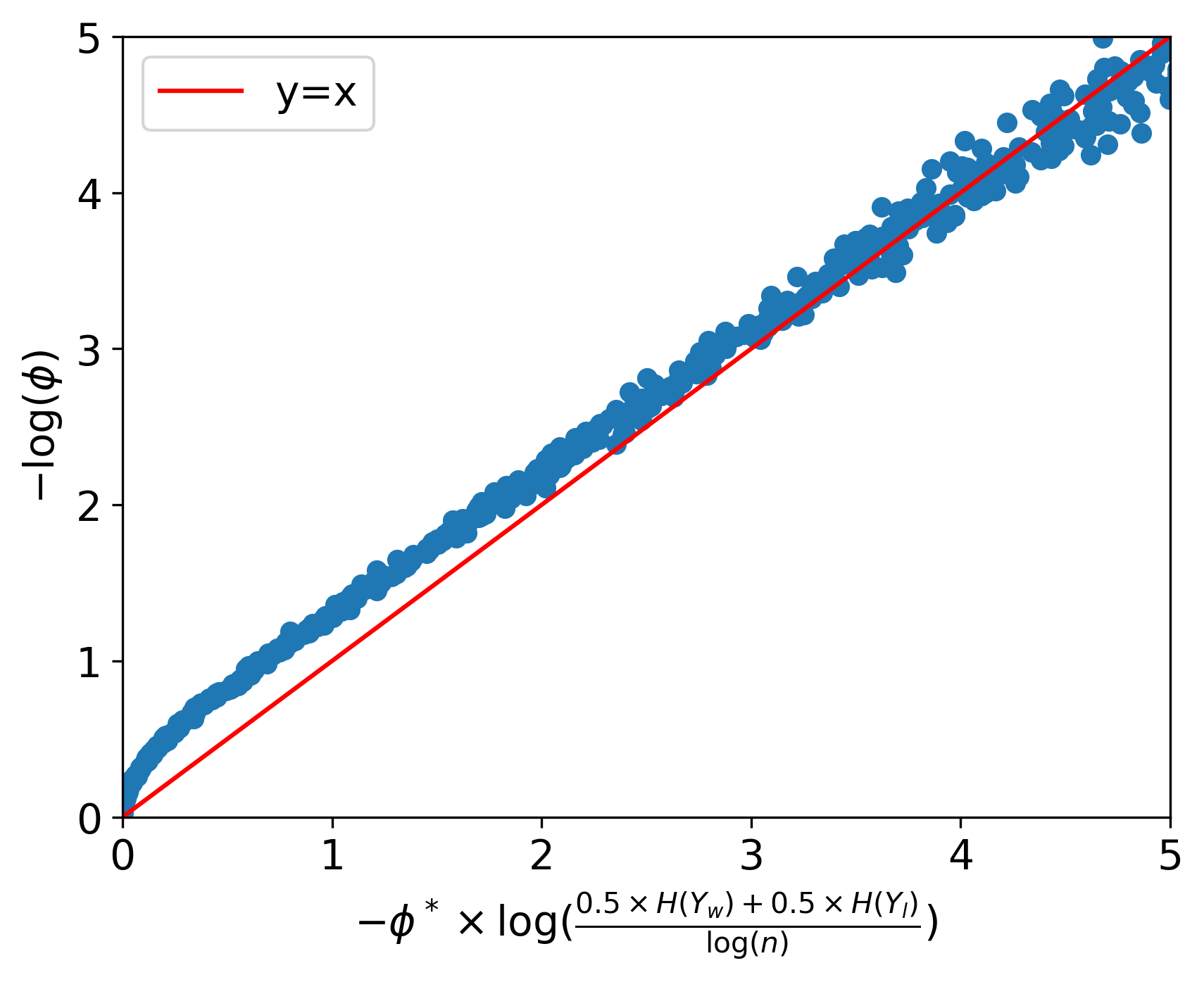}
    \caption{\small{Our proposed estimate matches the true (neg-log) dispersion under a Mallows model}.}
    \label{fig:Dispersion Matching}
  \end{minipage}
\end{figure}

The expression in (\ref{eq:mallows_theta_prob}) shows that in the Mallows-$\theta$ ranking, the representation of $p^*(\cdot \mid x)$, with a \textit{link function} $g_{\theta, \phi(x)}(\cdot)$, resembles a sigmoid function but differs in that it is scaled by the term $-2\log\phi(x)$, which reflects the distribution's dispersion. Compared to BT, where the distribution is exactly in the form of sigmoid, Mallows-$\theta$ allows for greater flexibility in controlling the spread of the distribution function. This is particularly important in language modeling, as the concept of dispersion provides insight into how diverse people's preferences are for different completions/responses. As $\phi(x) \to 0$, $p^*(\cdot \mid x)$ is getting closer to a step function (i.e., Dirac delta) (as shown in Fig.~\ref{fig:dist_function}), corresponding to cases where the prompt $x$ has a clear, standard answer. Conversely, as $\phi(x) \to 1$, $p^*(\cdot \mid x)$ approaches a constant value of 0.5 (i.e. uniform), indicating that any answer to the prompt $x$ is equally reasonable.

In \eqref{eq:mallows_theta_prob}, by setting $-2\log\phi(x)=1$, we recover the DPO in \eqref{DPO objective}. 
Comparing the objective of the DPO
with that of
MallowsPO-$\theta$,
the key difference is the presence of an extra term $-\log \phi(x)$,
which reflects the dispersion of the prompt $x$.
Thus, MallowsPO-$\theta$ can be viewed as a generalized version of DPO that incorporates prompt dispersion. 
To see the effect of this additional term:
When dispersion is high ($\phi(x) \approx 1$), the term $-2\log\phi(x)$ approaches 0, reducing the weight on preference pairs; 
when dispersion $\phi(x)$ decreases, $-2\log\phi(x)$ increases, assigning more weight to preference pairs.

\paragraph{MallowsPO-$\phi$.} 
The consistent pairwise probability can be derived for 
the Mallows-$\phi$ model.
As shown in \cite{mallows} (see also \citep{Bu14, MW22}), the probability that a random permutation $\mu$ from the Mallows-$\phi$ model with dispersion $\phi$ agrees with $\mu_0$ on $\{y_1,y_2\}$ is
\begin{equation}\label{eq:mallows_phi_prob}
 \mathbb{P}\left(\mu(y_1\mid x)<\mu\left(y_2\mid x\right)\right) = g_{\phi, \phi(x)}(\mu_0(y_2 \mid x) - \mu_0(y_1 \mid x)),
\end{equation}
where 
\begin{equation}
\label{Mallows phi}
g_{\phi, \phi(x)}(s):=
\begin{cases}
\frac{s+1}{1-\phi(x)^{s+1}}-\frac{s}{1-\phi(x)^{s}}, \quad &s>0,\\
1-\frac{-s+1}{1-\phi(x)^{-s+1}}-\frac{s}{1-\phi(x)^{-s}}, \quad &s<0.
\end{cases}
\end{equation}

Similarly, using the (consistent) pairwise probability (\ref{eq:mallows_phi_prob}) derived from the Mallows-$\phi$ model, we formulate our policy optimization objective.

\begin{theorem}[MallowsPO-$\phi$]
\label{thm:mallowspo_phi}
$~$
Given a prompt $x$, suppose the completions $y$ follow a Mallows preference distribution $p^*$ as in (\ref{eq:mallows_phi_prob}), i.e. $\mu$ is drawn from Mallows-$\phi$ with chosen $\phi(x)$ and central ranking $\mu_0$, then for any $y_1$ and $y_2$, the optimal RLHF policy satisfies $\pi_{\mu_0}(\cdot \mid x)$ satisfies $p^*\left(y_1\succ y_2\mid x\right) =  g_{\phi, \phi(x)}\left(\beta \log \frac{\pi\left(y_w \mid x\right)}{\pi_{\text {ref }}\left(y_w \mid x\right)}-\beta \log \frac{\pi\left(y_l \mid x\right)}{\pi_{\text {ref }}\left(y_l \mid x\right)}\right)$. 
Therefore, the MallowsPO-$\phi$ policy optimization objective is:

\begin{equation}
\label{Mallows DPO-phi objective}
\min_{\pi} -\mathbb{E}_{\left(x, y_w, y_l\right) \sim \mathcal{D}}\left[\log g_{\phi, \phi(x)}\left(\beta \log \frac{\pi\left(y_w \mid x\right)}{\pi_{\text {ref }}\left(y_w \mid x\right)}-\beta \log \frac{\pi\left(y_l \mid x\right)}{\pi_{\text {ref }}\left(y_l \mid x\right)}\right)\right].
\end{equation}

\end{theorem}


\vspace{-5pt}
By specifying the underlying ranking to Mallows-$\phi$, 
we get a different link function $g_{\phi, \phi(x)}$, which also contains the dispersion index $\phi(x)$, resulting in a new preference optimization objective. We include more discussions about unified perspectives of Mallows-$\theta$ and Mallows-$\phi$ in Appendix \ref{Mallows unified}.

\vspace{-5pt}

\subsection{How to choose the dispersion index $\phi(x)$?}

As the dispersion index $\phi(x)\in(0,1]$ is unknown, computation or estimation of it requires learning via neural networks or other algorithms \citep{meila2010exponential}.
Here, however, we propose a more direct approach to estimate $\phi(x)$ without any pretraining or learning.
The idea is to qualitatively relate $\phi(x)$ to the empirical output distribution of the pretrained model,
on which we propose an `easy-to-compute' proxy to the negative log dispersion $-\log(\phi(x))$ for each prompt $x$.

Suppose the preference follows the Mallows-$\phi$ model. 
There are two extreme cases:
When $-\log(\phi(x))\rightarrow \infty$, we have:
$
p^*\left(y_1\succ y_2\mid x\right) =
\begin{cases}
1, & \text{ if } \mu_0(y_1\mid x) <  \mu_0(y_2\mid x),\\
0, & \text{ if } \mu_0(y_1\mid x) >  \mu_0(y_2\mid x).
\end{cases}
$
Thus, the probability distribution of the next token will concentrate on a point mass. 
When $-\log(\phi(x))\rightarrow 0$, we have:
$p^*\left(y_1\succ y_2\mid x\right) = \frac{1}{2},$
so the next token will be uniformly distributed. 

The above observation motivates us to use Shannon's entropy $H(\cdot)$.
Note that $H(X) = 0$ when $X$ is a point mass,
and $H(X)= \log n$ when $X$ is uniform on $n$ points. 
Thus, we propose:
\begin{equation}
\label{dispersion estimator}
-\phi^*  \log\left(H(\pi(\cdot\mid x))/\log n\right),
\text{ for a given constant } \phi^*>0,
\end{equation}
as a proxy to $-\log \phi(x)$,
where $\pi(\cdot\mid x)$ can be either the pretrained LM model $\pi^{\mathrm{PRE}}$ or the SFT model $\pi^{\mathrm{SFT}}$. Furthermore, we approximate the entropy term in \eqref{dispersion estimator}
via a realization of a sequence of $N=\max(|Y^w|,|Y^l|)$ tokens $\{Y_i^w,Y_i^l\}_{i=1, \ldots, N}$ given the prompt $X$:
\begin{equation}
H(\pi(\cdot\mid X))\approx \frac{1}{2} \sum_{i=1}^{N-1}\left[H(Y_{i+1}\mid Y_i=Y^w_i)+H(Y_{i+1}\mid Y_i=Y^l_i)\right],
\end{equation}
which can be {\em easily} computed by the logits of the model given the output data. In this case, $n = k^N$, where $k$ is the token size. This is also related to the predictive entropy \citep{predictive_entropy,mackay1992information} of the next-token predictions. 

\textbf{Accuracy of the estimate.} To validate our proposed estimate \eqref{dispersion estimator}, we consider the similar `bandit' setup in \cite{tang2019mallows}. We draw rankings $\mu$ from a Mallows-$\phi$ ranking model, and then obtain a pair of winning/losing actions by choosing the highest/lowest ranked elements in the ranking $\mu$. 
We plot $-\phi^* \cdot \log \left(\frac{H\left(Y^w\right)+ H\left(Y^l\right)}{2 \log n}\right)$, given the preferences data ($x$-axis) and the true dispersion ($y$-axis) that these data are generated from. Figure \ref{fig:Dispersion Matching} shows that our proposed estimator indeed matches the true dispersion, which heuristically reflects the accuracy of our estimate.


\section{Perspectives on MallowsPO}
\label{sc4}
In this section, we provide several alternative perspectives on 
MallowsPO in Theorem \ref{thm:mallowspo_theta}--\ref{thm:mallowspo_phi},
with the proofs given in Appendix \ref{sec:proofs}.
We say a DPO is {\em directed by $g(\cdot)$} if the preference distribution can be expressed as 
$p^*\left(y_1 \succ y_2 \mid x\right) = g\left(r^*(x, y_1)-r^*\left(x, y_2\right)\right)$
for some reward function $r^*$.
Thus, Bradley-Terry based DPO is directed by the sigmoid function $\sigma(\cdot)$.

\subsection{Dispersion weighted objectives}
The following results show that MallowsPO can be viewed as a DPO with either the reward or the KL-regularizer weighted by the dispersion index.

\begin{proposition}[MallowsPO-$\theta$ as dispersion weighted DPO]
\label{thm:MallowsPO_theta_as_weigted_DPO}
$~$
Let $c(x)=-2 \log \phi(x)$. Then, MallowsPO-$\theta$
is the same as a DPO with either the reward weighted by $c(x)$ or the KL-regularizer weighted by $\beta c(x)$:
$\max _{\pi} \mathbb{E}_{x \sim \mathcal{D}}\left[\mathbb{E}_{y \sim \pi_\theta(y \mid x)}\left[c(x)^{-1}r^*(x, y)\right]-\beta \mathrm{KL}\left(\pi(\cdot\mid x) \| \pi_{\mathrm{ref}}(\cdot \mid x)\right)\right]$,
or 
$\max _{\pi} \mathbb{E}_{x \sim \mathcal{D}}\left[\mathbb{E}_{y \sim \pi_\theta(y \mid x)}\left[r^*(x, y)\right]- \beta c(x)  \mathrm{KL}\left(\pi(\cdot\mid x) \| \pi_{\mathrm{ref}}(\cdot \mid x)\right)\right]$.
\end{proposition}

\begin{proposition}[MallowsPO-$\phi$ as dispersion weighted DPO]
Denoting $\phi(x)=t$ in Eqn. \ref{Mallows phi} yields
\begin{equation}
\label{standard phi}
g(s):=\frac{1-\operatorname{sgn}(s)}{2}+\operatorname{sgn}(s)\left(
\frac{|s|+1}{1-t^{|s|+1}}-\frac{|s|}{1-t^{|s|}}\right).
\end{equation}
Let $c(x)=-2 \log \phi(x)$ as before. Then, MallowsPO-$\phi$
is the same as a DPO directed by $g(\cdot)$ as in \eqref{standard phi}, and with either the reward weighted by $c(x)$ or the KL-regularizer weighted by $\beta c(x)$:
$\max _{\pi} \mathbb{E}_{x \sim \mathcal{D}}\left[\mathbb{E}_{y \sim \pi_\theta(y \mid x)}\left[c(x)^{-1}r^*(x, y)\right]-\beta \mathrm{KL}\left(\pi(\cdot\mid x) \| \pi_{\mathrm{ref}}(\cdot \mid x)\right)\right]$, 
or
$\max _{\pi} \mathbb{E}_{x \sim \mathcal{D}}\left[\mathbb{E}_{y \sim \pi_\theta(y \mid x)}\left[r^*(x, y)\right]- \beta c(x)  \mathrm{KL}\left(\pi(\cdot\mid x) \| \pi_{\mathrm{ref}}(\cdot \mid x)\right)\right]$.
\end{proposition}

\subsection{Connection to $\Psi$PO}
The objective of $\Psi$PO \citep{IPO} is
$\max_{\pi}\mathbb{E}_{x \sim \mathcal{D}}[\mathbb{E}_{y \sim \pi(\cdot \mid x),\, y' \sim \tilde{\pi}(\cdot \mid x)}\left[\Psi\left( p^*( y \succ y' \mid x) \right)\right]- \beta \mathrm{KL}(\pi(\cdot\mid x) \| \pi_{\mathrm{ref}}(\cdot \mid x))]$,
where $\Psi: [0,1] \to \mathbb{R}$ is a non-decreasing function,
and $\tilde{\pi}(\cdot \mid x)$ is an arbitrary policy (referred to as the {\em behavior policy}).
It is readily verified that setting $\Psi(s) = \log\left(\frac{s}{1-s}\right)$ reduces $\Psi$PO to the Bradley-Terry based DPO.  Roughly speaking, the function $\Psi$ can be viewed as the inverse of the link function, $\Psi(\sigma(s)) = \log \left(\frac{\sigma(s)}{1 - \sigma(s)} \right) =s$. The question is whether MallowsPO can be reduced to $\Psi$PO for some suitably chosen $\Psi(\cdot)$. 
Assume such a function exists, which we denote as $\Psi^M(\cdot)$.  
From the Mallows-$\phi$ model in Theorem \ref{thm:mallowspo_phi}, we have
\begin{equation}
\begin{aligned}
\mathbb{E}_{y_2 \sim \bar{\pi}(\cdot\mid x)}\left[\Psi^{M}\left(p^*\left(y_1 \succ y_2\mid x\right)\right)\right] 
& =\mathbb{E}_{y_2 \sim \bar{\pi}(\cdot\mid x)}\left[\Psi^{M}\left(g_x(r(x, y_1)-r(x, y_2))\right)\right] \\
& \neq r(x, y_1)-\mathbb{E}_{y_2 \sim \bar{\pi}(\cdot\mid x)}\left[r\left(x, y_2\right)\right] ,
\end{aligned}
\end{equation}
i.e., for any $\Psi^{M}(\cdot)$ that is prompt-independent,
MallowsPO cannot be an instance of $\Psi$PO. This calls for extending $\Psi$PO to take into account prompt dispersion.

{\bf \em Generalized $\Psi$PO.} Let $\tilde{\Psi}(x, p)$ depend on the prompt $x$ as well as the preference distribution $p$. The
generalized $\Psi$PO takes the form:
\begin{equation}
\label{eq:generalpsi}
\max_{\pi}\mathbb{E}_{x \sim \mathcal{D}}\left[\mathbb{E}_{y \sim \pi_\theta(\cdot \mid x),\, y' \sim \tilde{\pi}(\cdot \mid x)}\left[\tilde{\Psi}\left(x, p^*( y \succ y' \mid x) \right)\right]-\beta \mathrm{KL}\left(\pi(\cdot\mid x) \| \pi_{\mathrm{ref}}(\cdot \mid x)\right)\right].
\end{equation}
A special instance is when $\tilde{\Psi}(x, p) = f(x) \Psi(p)$ is separable:
\begin{equation}
\label{eq:generalpsi2}
\max_{\pi}\mathbb{E}_{x \sim \mathcal{D}}\left[\mathbb{E}_{y \sim \pi_\theta(\cdot \mid x),\, y' \sim \tilde{\pi}(\cdot \mid x)}\left[f(x)\Psi\left( p^*( y \succ y' \mid x) \right)\right]-\beta \mathrm{KL}\left(\pi(\cdot\mid x) \| \pi_{\mathrm{ref}}(\cdot \mid x)\right)\right].
\end{equation}

\vspace{-5pt}
\begin{theorem}[MallowsPO as generalized $\Psi$PO]\label{thm:MallowsPO_as_psiPO}
(i) MallowsPO-$\theta$ 
(directed by $\sigma(\cdot)$) can be reduced to
the generalized $\Psi$PO in \eqref{eq:generalpsi2} with 
$\Psi(s)=\log \left(\frac{s}{1-s}\right)$ and $f(x)=-\frac{1}{2 \log \phi(x)}$.

(ii) MallowsPO-$\phi$ 
(directed by $g(\cdot)$) can be reduced to the generalized $\Psi$PO in \eqref{eq:generalpsi2} with $\Psi(s) = g^{-1}(s)$ and $f(x)=-\frac{1}{\log \phi(x)}$.
\end{theorem}

\section{Experiments}
\label{sc5}

In this section, we evaluate the capability of our proposed MallowsPO to learn the preferences in comparison with DPO.
First, we use the preferences dataset of IMDB \citep{IMDB} datasets 
and Anthropic Helpful and Harmless dialogue \citep{AnthropicHH} dataset to provide evidence that human preferences may be diversed. 
Next, we consider a synthetic bandit problem to demonstrate the effectiveness of our proposed MallowsPO-$\phi$, even without prompt dispersions. 
We further conduct experiments on tasks such as \textbf{conditional generation} (IMDB) and \textbf{ dialogue} (Anthropic HH, UltraFeedback). 
Our findings show that MallowsPO outperforms DPO with an evident margin,
both for in-distribution performance and out-of-distribution generalization capability. \footnote{Our code is available at ~\url{https://github.com/haoxian-chen/MallowsPO}.}

\subsection{Evidence of preference dispersion}
\label{sc51}
A first natural question is: are human preferences dispersed? 
To verify this key motivation for our work, we plot the distribution of the dispersion estimators given the SFT model and pairwise preferences. 
Recall from Section \ref{sc3} that the dispersion estimator is:
\vspace{-5 pt}
\begin{equation}
\label{eqn:dispersion estimator final}
-\phi^* \log(\frac{1}{2 N\log k} \sum_{i=1}^{N-1}\left[H(Y_{i+1}\mid Y_i=y^w_i)+H(Y_{i+1}\mid Y_i=y^l_i)\right]),
\vspace{-5 pt}
\end{equation}
and we take the hyperparameter $\phi^*>0$ such that the empirical mean is equal to 1 (as in DPO), so we \textbf{do not} need to tune this scaling constant. 
Note that this scaling results in our final estimate acting as a relative dispersion level compared to the whole dataset: 
when a prompt's dispersion parameter is large, i.e., close to 1, \eqref{eqn:dispersion estimator final} will be smaller than 1. 
In contrast, \eqref{eqn:dispersion estimator final} will be much larger than 1 if the prompt's dispersion parameter is close to 0 (or there is less disagreement about the answer to the prompt). 
We formally call this {\it neg-log dispersion estimate} throughout the rest of the paper.

\begin{figure}[!htbp]
  \centering
  \begin{minipage}{0.49\linewidth}
    \centering
    \includegraphics[width=0.85\linewidth]{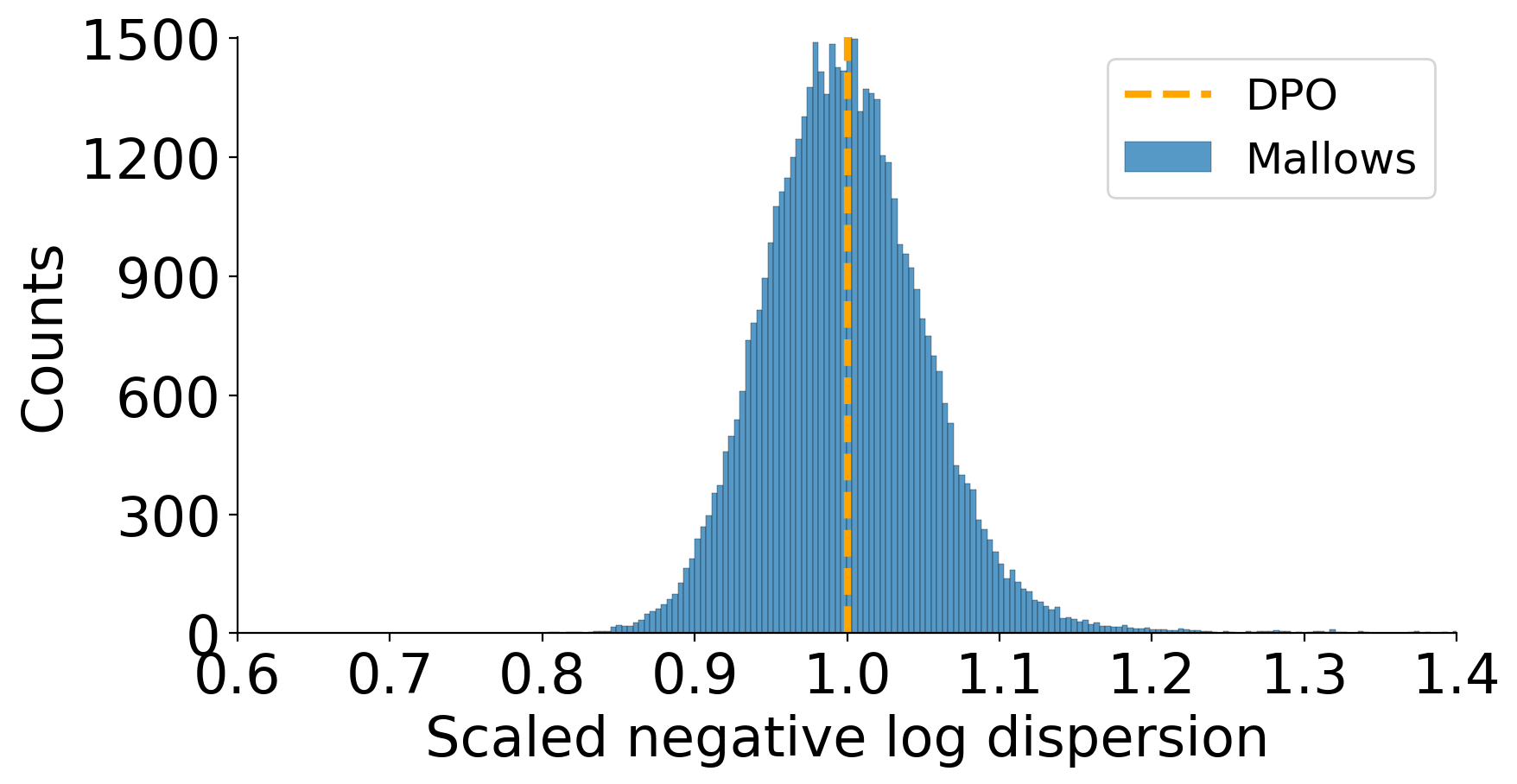}
    \subcaption{\scriptsize{IMDB preference dispersion distribution.}}
    \label{fig:imdb SFT dispersion estimator distribution}
  \end{minipage} 
  \begin{minipage}{0.49\linewidth}
    \centering
    \includegraphics[width=0.85\linewidth]{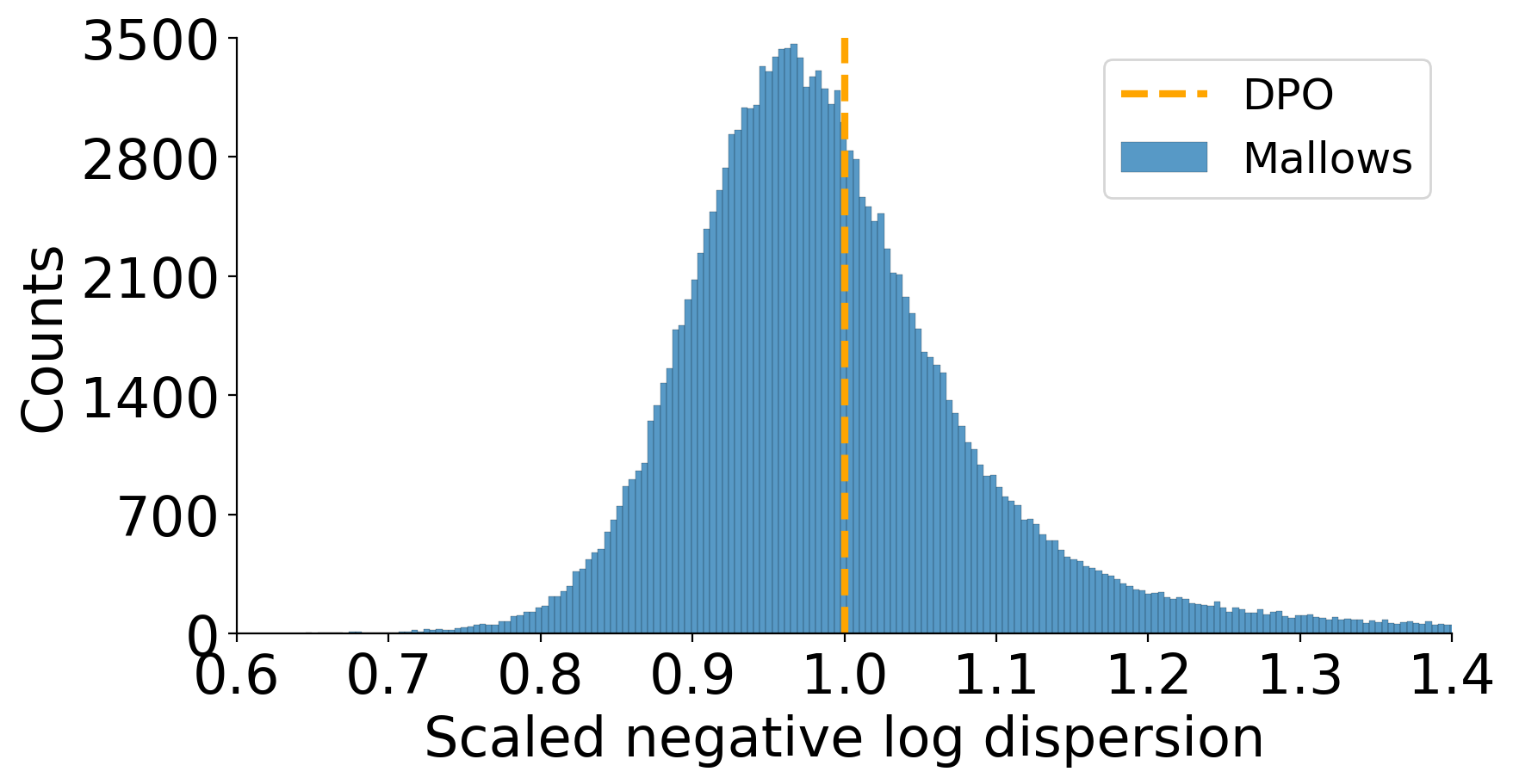}
    \subcaption{\scriptsize{Anthropic-HH preference dispersion distribution.}}
    \label{fig:hh SFT dispersion estimator distribution}
  \end{minipage}
\end{figure}

We find that for the task of conditional generation such as IMDB, its human preferences (Fig. \ref{fig:imdb SFT dispersion estimator distribution}) are not quite diverse: the neg-log dispersion estimates are located near $1$, and almost all the estimates range from $0.8$ to $1.2$.
However, for tasks such as single dialogue, Fig. \ref{fig:hh SFT dispersion estimator distribution} shows that human preferences are relatively more dispersed: the distribution is both skewed and of higher variance. 
As shown in Figure \ref{fig:prompt dispersion with examples}, prompts with high dispersion or those that will lead to human disagreement on preferences indeed have a neg-log dispersion estimate smaller than $1$, while those with low dispersion have the neg-log dispersion estimate located at the right-hand side (larger than 1).
More examples with low/high dispersion are provided in Appendix \ref{app: more examples for prompts with low or high dispersion}.

\subsection{MallowsPO-$\phi$ mitigates reward collapse}\label{sec:bandit}
We study MallowsPO in a synthetic bandit experiment
with no contextual information $x$,
and compare it with DPO to test their ability to produce diversified policies and avoid reward collapse.
Moreover, we operate under the constraint of having a limited number of observations.
There are two reasons to explore this setting.
First, the bandit facilitates a clear analysis without introducing the complication of the context $x$.
Second, the limited data availability tests the ability of the approaches to produce diversified policies and avoid reward collapse.

Concretely, we consider five arms, each associated with a random reward drawn from a probability distribution.
Preference between any two picked arms is determined by the random reward realizations, with larger reward being preferred. 
In the experiment, we collect $16$ pairwise observations,
and evaluate the performance of different approaches by computing the efficient frontiers 
(1) across different parameters $\beta$, and (2) across different epochs. 
The details are provided in Appendix B.1.

\begin{figure}[!htbp]
  \centering
    \includegraphics[width=0.76\linewidth]{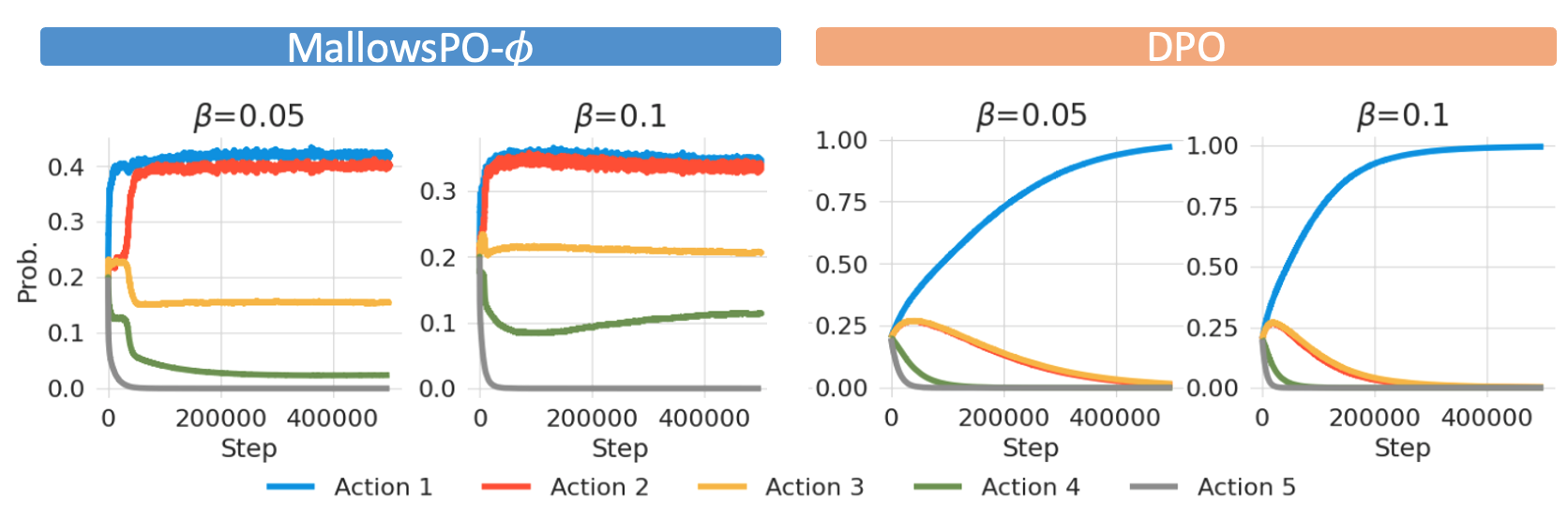}
    \caption{\small{Training curves of MallowsPO-$\phi$ and DPO for $\beta=0.05$ and $\beta=0.1$.}}
    \label{fig:bandit_training_curves}
\end{figure}

Figure \ref{fig:bandit_combined} displays the efficient frontiers for MallowsPO-$\phi$ and DPO. 
Figure \ref{fig:bandit_betas} shows that MallowsPO-$\phi$ has a more efficient frontier:
(1) With the same KL divergence, MallowsPO-$\phi$ achieves a higher reward, especially when $\beta$ is small.
(2) Over all possible $\beta$, the best reward that MallowsPO-$\phi$ achieves is higher than that of DPO. 
(3) MallowsPO-$\phi$ avoids reward collapse as $\beta$ gets smaller.
That is, MallowsPO-$\phi$ assigns a certain probability to the potentially good arms, 
as opposed to DPO that tends to assign only to the ``best'' arm (see Figure \ref{fig:bandit_training_curves}).
Figure \ref{fig:bandit_epochs} shows that during the training process,
MallowsPO-$\phi$ leads to the policies that have both high rewards and small KL divergence. 

\begin{figure}[!htbp]
  \centering
  \begin{subfigure}[b]{0.33\textwidth}
    \centering
    \includegraphics[width=\linewidth]{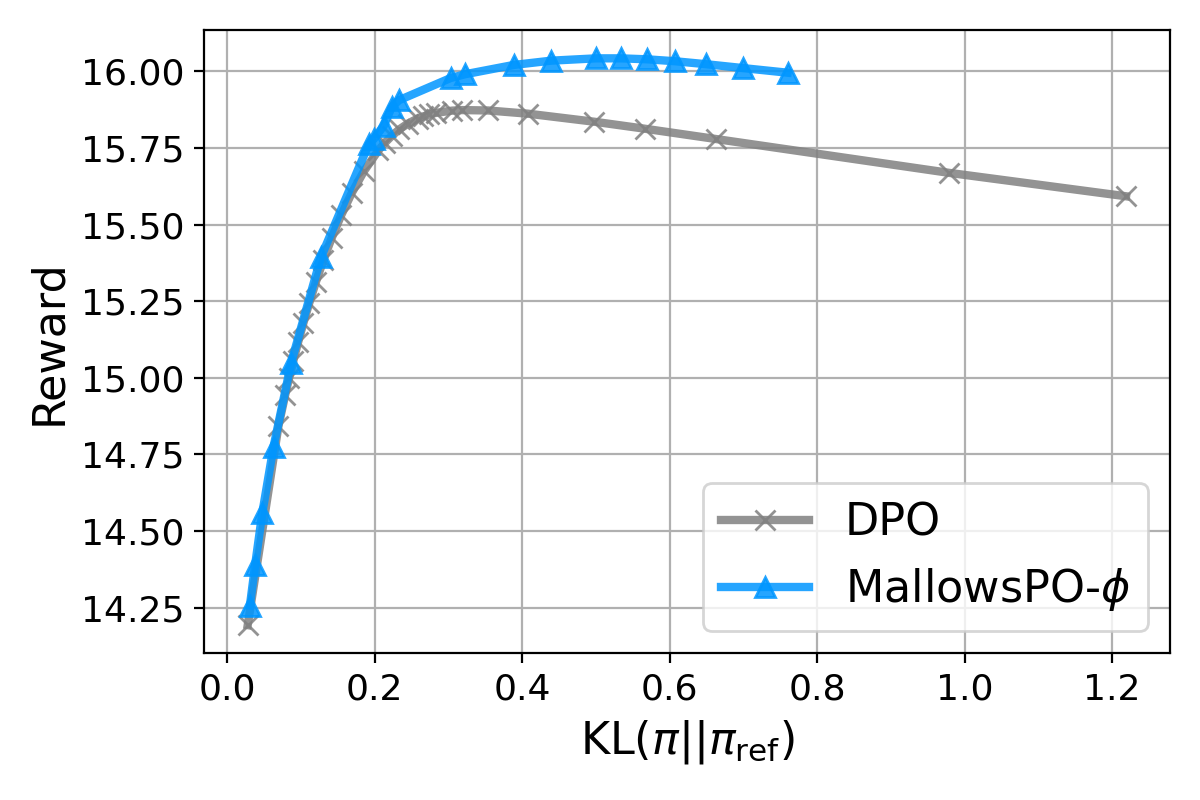}
    \subcaption{}
    \label{fig:bandit_betas}
  \end{subfigure}%
  \begin{subfigure}[b]{0.33\textwidth}
    \centering
    \includegraphics[width=0.99\linewidth]{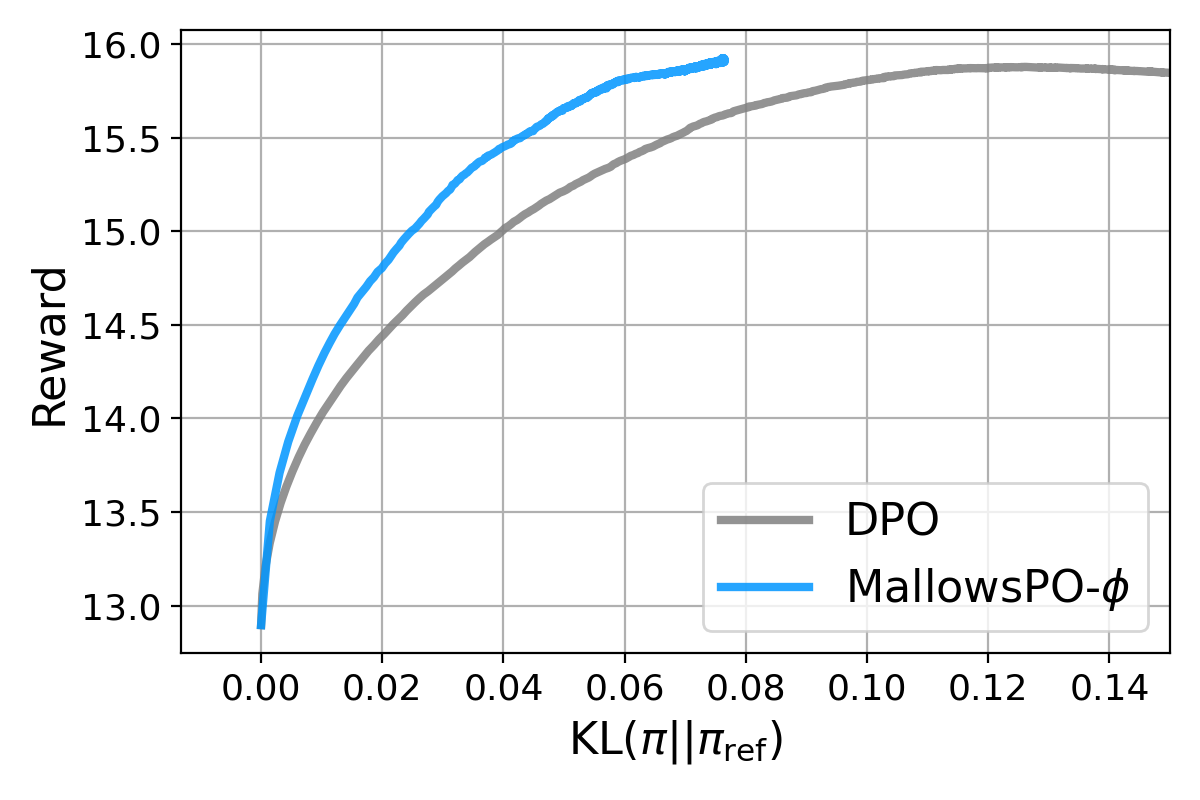}
    \subcaption{}
    \label{fig:bandit_epochs}
  \end{subfigure}
  \begin{subfigure}[b]{0.33\textwidth}
    \centering
    \includegraphics[width=0.87\linewidth]{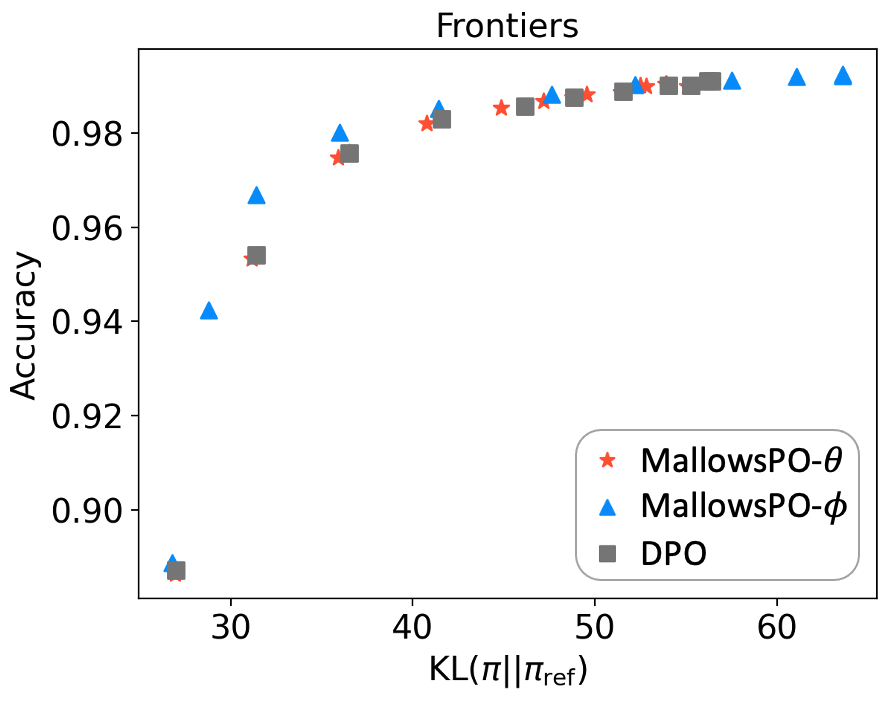}
    \subcaption{}
    \label{fig:EF of models}
  \end{subfigure}
  \caption{\small{\textbf{(a)}. Reward vs KL for the policy with different $\beta$'s. \textbf{(b)}. Reward vs KL every $100$ epochs, averaging over the four policies with $\beta \in \{0.05, 0.1, 0.5, 1.0\}$. \textbf{(c)} Accuracy vs KL achieved by MallowsPO and DPO.}}
  \label{fig:bandit_combined}
\end{figure}
\vspace{-8pt}

\subsection{MallowsPO yields better tradeoff between accuracy and regularization}
In the task of conditional generation for IMDB, $x$ is a prefix of movie review, and LM is to generate output $y$ with positive sentiment.
Following the setting in \cite{DPO}, 
we first fine-tune GPT-2-large on the training split of IMDB datasets until convergence to get the SFT model, and use the pairwise preference data from \cite{f-DPO} to further fine-tune it by DPO and MallowsPO.


Figure \ref{fig:EF of models} displays the efficient frontiers (during the training process) for DPO and MallowsPO.
We observe that the performances of MallowsPO-$\theta$ and DPO are close. The similarity is likely due to the nature of the task -- controllable comment generation, which is expected to exhibit smaller dispersion, as evidenced in Figure \ref{fig:imdb SFT dispersion estimator distribution}.
MallowsPO-$\phi$ outperforms both, achieving the same accuracy (evaluated by the reward model) at a smaller KL divergence to the SFT model/policy.

\subsection{MallowsPO enhances both in/out-of distribution performances}\label{sec:id_and_ood_results}
\begin{figure}[!htbp]
  \centering
  \includegraphics[width=0.9\linewidth]{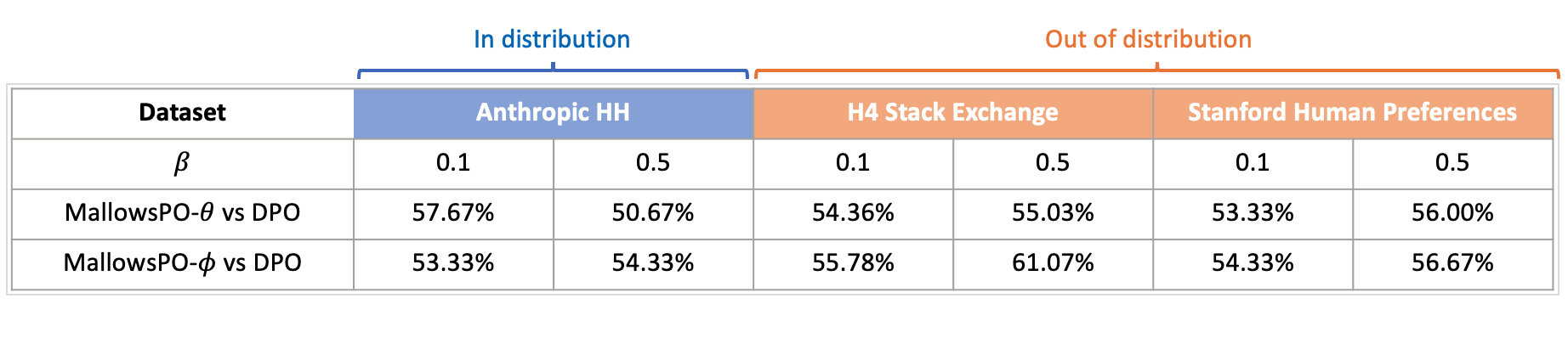}
  \vspace{-5pt}
  \caption{\small{Win rates computed by GPT-4 for responses on both the in- and out-of distribution dataset.}}
  \label{fig: LM results}
\end{figure}

We compare the performances of MallowsPO and DPO in terms of the win rate evaluated by GPT4, and generalization capability on the out-of-distribution datasets.
In the experiment, we choose $\beta$ to be $0.1$ and $0.5$ 
since it has been observed \citep{kirk2023understanding} that increased $\beta$ value leads to a drop both in performance and per-input diversity of RLHF and DPO. Results are shown in Figure \ref{fig: LM results}.

For the {\em in-distribution test}, we first fine-tune a pretrained Pythia-2.8B model on the training set of Anthropic HH dataset using MallowsPO and DPO,
and then evaluate their responses on a subset of its test split.
GPT-4 serves as the evaluator, and compares pairs of responses.
We observe that MallowsPO has an edge over DPO.
For the {\em out-of-distribution test},
we apply the models, fine-tuned on the train split of the Antropic HH dataset, to other datasets with different input distributions.
The H4 Stack Exchange Preferences Dataset (SE) \citep{se}
and Stanford Human Preferences (SHP)  \citep{shp}
are used for evaluation.
The advantage of dispersion on generalization becomes apparent, 
as MallowsPO shows more improvement compared to the in-distribution case. 

We also compare MallowsPO-$\theta$ with DPO in fine-tuning the Pythia-2.8B model, 
with ArmoRM \citep{wang2024interpretable} serving as the evaluator.
The result indicates that MallowsPO-$\theta$ achieves consistently higher win rates than DPO across all cases, with an impressive win rate of around 70\% in the in-distribution test.
Details are provided in Appendix \ref{append:Armo}. 

\vspace{-5pt}

\subsection{MallowsPO enhances SOTA Llama3-8B-Instruct Models}\label{exp:llama3}
We illustrate the scalability of our method through experiments on fine-tuning Llama3-8B-Instruct Model on UltraFeedback Dataset. We follow the same setup in RLHFlow \citep{dong2024rlhf} and SimPO \citep{SimPO}, 
as we generate five answers from Llama3-8B-Instruct for each prompt in UltraFeedback, rank them with scores evaluated by ArmoRM \citep{wang2024interpretable},
and choose the best/worst one as winning/losing answer to form the preference datasets. For a fair comparison, we compare MallowsPO with DPO, using different hyperparameters: $\beta$ and learning rate $lr$
for the task of Alpaca Eval V2.
The results are shown in Appendix \ref{sec:ablation}:

\begin{table}[htbp!]
\centering
\small
\begin{tabular}{cc : cc : cc}
\toprule
$\beta$ & $lr$ & \multicolumn{2}{c:}{LC Win Rate} & \multicolumn{2}{c}{Win Rate} \\
\midrule
& & DPO & MallowsPO & DPO & MallowsPO \\
0.01 & $5e^{-7}$  & 42.55\% (0.79) & \textbf{43.10\%} (0.77)& 42.02\% (1.53) & \textbf{43.02\%} (1.57) \\
\midrule
& & IPO & MallowsIPO & IPO & MallowsIPO \\
0.005 & $1e^{-6}$  & 43.38\% (0.84) & \textbf{44.73\%} (0.87)  & 43.52\% (1.45) & \textbf{44.87\%} (1.46) \\
\midrule
& & SimPO & MallowsSimPO & SimPO & MallowsSimPO \\
10 & $1e^{-6}$  & 50.04\% (0.77) & \textbf{51.89\%} (0.81) & 42.11\% (1.46) & \textbf{43.76\%} (1.47) \\
\bottomrule
\end{tabular}
\caption{\small{Win rate comparison between SOTA fine-tuning methods and their enhanced versions using our MallowsPO as a plugin with optimized $\beta$ and $lr$. Standard deviations are right next to the reported metric.}}
\label{tab:misc_win_rates}
\end{table}

\begin{wrapfigure}{r}{0.45\textwidth}
\vspace{-5pt}
  \centering
  \includegraphics[width=\linewidth]{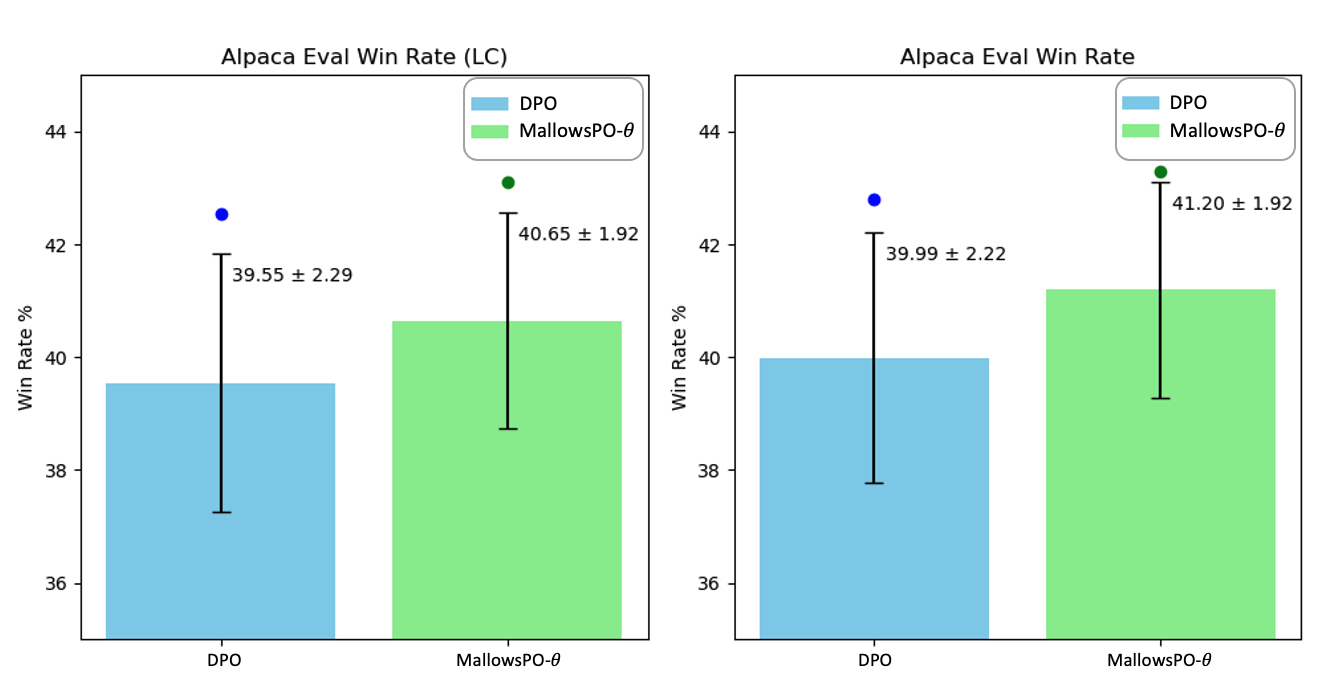}
  \caption{\small{Win rates computed by GPT-4 for responses on Alpaca Eval V2.}}
  \label{fig: Alpaca results}

\vspace{-10pt}
\end{wrapfigure}

When $\beta=0.01$ and $lr=5e^{-7}$, for which DPO and MallowsPO both achieve the best performance, we used 10 random seeds to show the statistical significance:  MallowsPO outperforms DPO both in mean or the best performance across random seeds, and also has smaller variance (see Figure \ref{fig: Alpaca results}).

We also adapt the idea of dispersion index in MallowsPO to IPO and SimPO, leading to MallowsIPO and MallowsSimPO.
As shown in Table \ref{tab:misc_win_rates}, both MallowsIPO and MallowsSimPO beat their vanilla counterparts 
(using hyperparameters proposed in \cite{IPO} and \cite{SimPO}).

\section{Conclusion}
\label{sc6}
We have developed in this paper a novel approach, the MallowsPO, to fine-tune LLM. 
A distinct feature of this approach is a dispersion index, 
which naturally captures the dispersion of human preference to prompts, and can be systematically incorporated into the reward function as a weight factor, thus ushering in a new class of dispersion-weighted DPO models.
We demonstrate empirically how MallowsPO achieves improved performance in a broad array of benchmark tasks, including synthetic bandit selection, controllable generation, and dialogues. The effectiveness holds for both small and large representative models, such as Pythia 2.8B and Llama3-8B-Instruct. 

Furthermore, both curriculum learning and personalization alignment through contextual scaling present promising future directions and can be expanded based on our work (see Appendix \ref{app: future research}).

There are a few issues that we have yet to address in this study, for instance, to explore why MallowsPO outperforms DPO, how the dispersion index contributes to performance improvement, what guidelines to follow to set the $\beta$ value, and how dispersion index can be more accurately approximated. These will be pursued in our future work.

\section*{Acknowledgement}

Hanyang Zhao, Haoxian Chen and Wenpin Tang are supported by NSF grant DMS-2206038.
Wenpin Tang acknowledges financial support by the Tang Family Assistant Professorship.
The works of Haoxian Chen, Hanyang Zhao, Henry Lam
and David Yao are part of a Columbia-CityU/HK collaborative project that is supported by InnotHK
Initiative, The Government of the HKSAR and the AIFT Lab. 
Additionally, Haoxian Chen is supported by the Amazon CAIT fellowship.
Henry Lam and Wenpin Tang receive support from the Columbia Innovation Hub grant.

\bibliography{iclr2025_conference}

\begin{thebibliography}{56}
\providecommand{\natexlab}[1]{#1}
\providecommand{\url}[1]{\texttt{#1}}
\expandafter\ifx\csname urlstyle\endcsname\relax
  \providecommand{\doi}[1]{doi: #1}\else
  \providecommand{\doi}{doi: \begingroup \urlstyle{rm}\Url}\fi

\bibitem[Achiam et~al.(2023)Achiam, Adler, Agarwal, Ahmad, Akkaya, Aleman, Almeida, Altenschmidt, Altman, and Anadkat]{GPT4}
Josh Achiam, Steven Adler, Sandhini Agarwal, Lama Ahmad, Ilge Akkaya, Florencia~Leoni Aleman, Diogo Almeida, Janko Altenschmidt, Sam Altman, and Shyamal Anadkat.
\newblock G{P}{T}-4 technical report.
\newblock 2023.
\newblock arXiv:2303.08774.

\bibitem[Amini et~al.(2024)Amini, Vieira, and Cotterell]{ODPO}
Afra Amini, Tim Vieira, and Ryan Cotterell.
\newblock Direct preference optimization with an offset.
\newblock 2024.
\newblock arXiv:2402.10571.

\bibitem[Azar et~al.(2024)Azar, Guo, Piot, Munos, Rowland, Valko, and Calandriello]{IPO}
Mohammad~Gheshlaghi Azar, Zhaohan~Daniel Guo, Bilal Piot, Remi Munos, Mark Rowland, Michal Valko, and Daniele Calandriello.
\newblock A general theoretical paradigm to understand learning from human preferences.
\newblock In \emph{AISTATS}, pp.\  4447--4455, 2024.

\bibitem[Bai et~al.(2022{\natexlab{a}})Bai, Jones, Ndousse, Askell, Chen, DasSarma, Drain, Fort, Ganguli, and Henighan]{AnthropicHH}
Yuntao Bai, Andy Jones, Kamal Ndousse, Amanda Askell, Anna Chen, Nova DasSarma, Dawn Drain, Stanislav Fort, Deep Ganguli, and Tom Henighan.
\newblock Training a helpful and harmless assistant with reinforcement learning from human feedback.
\newblock 2022{\natexlab{a}}.
\newblock arXiv:2204.05862.

\bibitem[Bai et~al.(2022{\natexlab{b}})Bai, Kadavath, Kundu, Askell, Kernion, Jones, Chen, Goldie, Mirhoseini, and McKinnon]{bai2022constitutional}
Yuntao Bai, Saurav Kadavath, Sandipan Kundu, Amanda Askell, Jackson Kernion, Andy Jones, Anna Chen, Anna Goldie, Azalia Mirhoseini, and Cameron McKinnon.
\newblock Constitutional {A}{I}: Harmlessness from {A}{I} feedback.
\newblock 2022{\natexlab{b}}.
\newblock arXiv:2212.08073.

\bibitem[Bengio et~al.(2009)Bengio, Louradour, Collobert, and Weston]{bengio2009curriculum}
Yoshua Bengio, J{\'e}r{\^o}me Louradour, Ronan Collobert, and Jason Weston.
\newblock Curriculum learning.
\newblock In \emph{Proceedings of the 26th annual international conference on machine learning}, pp.\  41--48, 2009.

\bibitem[Bradley \& Terry(1952)Bradley and Terry]{BT}
Ralph~Allan Bradley and Milton~E Terry.
\newblock Rank analysis of incomplete block designs: I. the method of paired comparisons.
\newblock \emph{Biometrika}, 39\penalty0 (3/4):\penalty0 324--345, 1952.

\bibitem[Brown et~al.(2020)Brown, Mann, Ryder, Subbiah, Kaplan, Dhariwal, Neelakantan, Shyam, Sastry, and Askell]{GPT3}
Tom Brown, Benjamin Mann, Nick Ryder, Melanie Subbiah, Jared~D Kaplan, Prafulla Dhariwal, Arvind Neelakantan, Pranav Shyam, Girish Sastry, and Amanda Askell.
\newblock Language models are few-shot learners.
\newblock In \emph{Neurips}, volume~33, pp.\  1877--1901, 2020.

\bibitem[Busa-Fekete et~al.(2014)Busa-Fekete, H{\"u}llermeier, and Sz{\"o}r{\'e}nyi]{Bu14}
R{\'o}bert Busa-Fekete, Eyke H{\"u}llermeier, and Bal{\'a}zs Sz{\"o}r{\'e}nyi.
\newblock Preference-based rank elicitation using statistical models: The case of {M}allows.
\newblock In \emph{ICML}, pp.\  1071--1079, 2014.

\bibitem[Calandriello et~al.(2024)Calandriello, Guo, Munos, Rowland, Tang, Pires, Richemond, Lan, Valko, and Liu]{calandriello2024human}
Daniele Calandriello, Daniel Guo, Remi Munos, Mark Rowland, Yunhao Tang, Bernardo~Avila Pires, Pierre~Harvey Richemond, Charline~Le Lan, Michal Valko, and Tianqi Liu.
\newblock Human alignment of large language models through online preference optimisation.
\newblock 2024.
\newblock arXiv:2403.08635.

\bibitem[Chan et~al.(2022)Chan, Ni, Ross, Vijayanarasimhan, Myers, and Canny]{Chan22}
David~M Chan, Yiming Ni, David~A Ross, Sudheendra Vijayanarasimhan, Austin Myers, and John Canny.
\newblock Distribution aware metrics for conditional natural language generation.
\newblock 2022.
\newblock arXiv:2209.07518.

\bibitem[Chen et~al.(2024{\natexlab{a}})Chen, Zhang, Luo, Chai, and Liu]{chen2024pad}
Ruizhe Chen, Xiaotian Zhang, Meng Luo, Wenhao Chai, and Zuozhu Liu.
\newblock Pad: Personalized alignment at decoding-time.
\newblock \emph{arXiv preprint arXiv:2410.04070}, 2024{\natexlab{a}}.

\bibitem[Chen et~al.(2024{\natexlab{b}})Chen, Deng, Yuan, Ji, and Gu]{chen2024self}
Zixiang Chen, Yihe Deng, Huizhuo Yuan, Kaixuan Ji, and Quanquan Gu.
\newblock Self-play fine-tuning converts weak language models to strong language models.
\newblock 2024{\natexlab{b}}.
\newblock arXiv:2401.01335.

\bibitem[Critchlow(1985)]{critchlow2012metric}
Douglas Critchlow.
\newblock \emph{Metric methods for analyzing partially ranked data}, volume~34.
\newblock Lecture notes in Statistics, Springer, 1985.

\bibitem[Diaconis(1988)]{diaconis1988group}
Persi Diaconis.
\newblock \emph{Group representations in probability and statistics}, volume~11.
\newblock Lecture Notes-Monograph Series, 1988.

\bibitem[Diaconis(1989)]{diaconis1989generalization}
Persi Diaconis.
\newblock A generalization of spectral analysis with application to ranked data.
\newblock \emph{Ann. Stat.}, pp.\  949--979, 1989.

\bibitem[Dong et~al.(2024)Dong, Xiong, Pang, Wang, Zhao, Zhou, Jiang, Sahoo, Xiong, and Zhang]{dong2024rlhf}
Hanze Dong, Wei Xiong, Bo~Pang, Haoxiang Wang, Han Zhao, Yingbo Zhou, Nan Jiang, Doyen Sahoo, Caiming Xiong, and Tong Zhang.
\newblock Rlhf workflow: From reward modeling to online rlhf.
\newblock 2024.
\newblock arXiv:2405.07863.

\bibitem[Dubey et~al.(2024)Dubey, Jauhri, Pandey, Kadian, Al-Dahle, Letman, Mathur, Schelten, Yang, and Fan]{Llama3}
Abhimanyu Dubey, Abhinav Jauhri, Abhinav Pandey, Abhishek Kadian, Ahmad Al-Dahle, Aiesha Letman, Akhil Mathur, Alan Schelten, Amy Yang, and Angela Fan.
\newblock The llama 3 herd of models.
\newblock \emph{arXiv preprint arXiv:2407.21783}, 2024.

\bibitem[Dubois et~al.(2024)Dubois, Li, Taori, Zhang, Gulrajani, Ba, Guestrin, Liang, and Hashimoto]{Alpacafarm}
Yann Dubois, Chen~Xuechen Li, Rohan Taori, Tianyi Zhang, Ishaan Gulrajani, Jimmy Ba, Carlos Guestrin, Percy~S Liang, and Tatsunori~B Hashimoto.
\newblock Alpacafarm: A simulation framework for methods that learn from human feedback.
\newblock In \emph{Neurips}, volume~36, 2024.

\bibitem[Ethayarajh et~al.(2022)Ethayarajh, Choi, and Swayamdipta]{shp}
Kawin Ethayarajh, Yejin Choi, and Swabha Swayamdipta.
\newblock Understanding dataset difficulty with $\mathcal{V}$-usable information.
\newblock In \emph{ICML}, pp.\  5988--6008, 2022.

\bibitem[Ethayarajh et~al.(2023)Ethayarajh, Xu, Jurafsky, and Kiela]{KTO}
Kawin Ethayarajh, Winnie Xu, Dan Jurafsky, and Douwe Kiela.
\newblock Human-aware loss functions (halos).
\newblock Technical report, Contextual AI, 2023.
\newblock https://github.com/ContextualAI/HALOs/blob/main/assets/report.pdf.

\bibitem[Fu et~al.(2022)Fu, Zhao, Tao, Wen, and Yan]{fu2022there}
Tingchen Fu, Xueliang Zhao, Chongyang Tao, Ji-Rong Wen, and Rui Yan.
\newblock There are a thousand hamlets in a thousand people's eyes: Enhancing knowledge-grounded dialogue with personal memory.
\newblock 2022.
\newblock arXiv:2204.02624.

\bibitem[Graves et~al.(2017)Graves, Bellemare, Menick, Munos, and Kavukcuoglu]{graves2017automated}
Alex Graves, Marc~G Bellemare, Jacob Menick, Remi Munos, and Koray Kavukcuoglu.
\newblock Automated curriculum learning for neural networks.
\newblock In \emph{international conference on machine learning}, pp.\  1311--1320. Pmlr, 2017.

\bibitem[Hern\'{a}ndez-Lobato et~al.(2014)Hern\'{a}ndez-Lobato, Hoffman, and Ghahramani]{predictive_entropy}
Jos\'{e}~Miguel Hern\'{a}ndez-Lobato, Matthew~W Hoffman, and Zoubin Ghahramani.
\newblock Predictive entropy search for efficient global optimization of black-box functions.
\newblock In \emph{NIPS}, volume~27, pp.\  918--926, 2014.

\bibitem[Hong et~al.(2024)Hong, Lee, and Thorne]{ORPO}
Jiwoo Hong, Noah Lee, and James Thorne.
\newblock Orpo: Monolithic preference optimization without reference model.
\newblock 2024.
\newblock arXiv:2403.07691.

\bibitem[Ivison et~al.(2023)Ivison, Wang, Pyatkin, Lambert, Peters, Dasigi, Jang, Wadden, Smith, and Beltagy]{Tulu2}
Hamish Ivison, Yizhong Wang, Valentina Pyatkin, Nathan Lambert, Matthew Peters, Pradeep Dasigi, Joel Jang, David Wadden, Noah~A Smith, and Iz~Beltagy.
\newblock Camels in a changing climate: Enhancing {L}{M} adaptation with {T}ulu 2.
\newblock 2023.
\newblock arXiv:2311.10702.

\bibitem[Kirk et~al.(2023)Kirk, Mediratta, Nalmpantis, Luketina, Hambro, Grefenstette, and Raileanu]{kirk2023understanding}
Robert Kirk, Ishita Mediratta, Christoforos Nalmpantis, Jelena Luketina, Eric Hambro, Edward Grefenstette, and Roberta Raileanu.
\newblock Understanding the effects of {R}{L}{H}{F} on {L}{L}{M} generalisation and diversity.
\newblock 2023.
\newblock arXiv:2310.06452.

\bibitem[Lambert et~al.(2023)Lambert, Tunstall, Rajani, and Thrush]{se}
Nathan Lambert, Lewis Tunstall, Nazneen Rajani, and Tristan Thrush.
\newblock Huggingface h4 stack exchange preference dataset, 2023.
\newblock URL \url{https://huggingface.co/datasets/HuggingFaceH4/stack-exchange-preferences}.

\bibitem[Lee et~al.(2023)Lee, Phatale, Mansoor, Lu, Mesnard, Bishop, Carbune, and Rastogi]{lee2023rlaif}
Harrison Lee, Samrat Phatale, Hassan Mansoor, Kellie Lu, Thomas Mesnard, Colton Bishop, Victor Carbune, and Abhinav Rastogi.
\newblock Rlaif: Scaling reinforcement learning from human feedback with {A}{I} feedback.
\newblock 2023.
\newblock arXiv:2309.00267.

\bibitem[Li et~al.(2016)Li, Monroe, Ritter, Galley, Gao, and Jurafsky]{li2016deep}
Jiwei Li, Will Monroe, Alan Ritter, Michel Galley, Jianfeng Gao, and Dan Jurafsky.
\newblock Deep reinforcement learning for dialogue generation.
\newblock 2016.
\newblock arXiv:1606.01541.

\bibitem[Maas et~al.(2011)Maas, Daly, Pham, Huang, Ng, and Potts]{IMDB}
Andrew Maas, Raymond~E Daly, Peter~T Pham, Dan Huang, Andrew~Y Ng, and Christopher Potts.
\newblock Learning word vectors for sentiment analysis.
\newblock In \emph{ACL}, pp.\  142--150, 2011.

\bibitem[MacKay(1992)]{mackay1992information}
David~JC MacKay.
\newblock Information-based objective functions for active data selection.
\newblock \emph{Neural computation}, 4\penalty0 (4):\penalty0 590--604, 1992.

\bibitem[Mallows(1957)]{mallows}
Colin~L Mallows.
\newblock Non-null ranking models. {I}.
\newblock \emph{Biometrika}, 44\penalty0 (1/2):\penalty0 114--130, 1957.

\bibitem[Mao \& Wu(2022)Mao and Wu]{MW22}
Cheng Mao and Yihong Wu.
\newblock Learning mixtures of permutations: groups of pairwise comparisons and combinatorial method of moments.
\newblock \emph{Ann. Statist.}, 50\penalty0 (4):\penalty0 2231--2255, 2022.

\bibitem[Meila \& Bao(2010)Meila and Bao]{meila2010exponential}
Marina Meila and Le~Bao.
\newblock An exponential model for infinite rankings.
\newblock \emph{J. Mach. Learn. Res.}, 11:\penalty0 3481--3518, 2010.

\bibitem[Meng et~al.(2024)Meng, Xia, and Chen]{SimPO}
Yu~Meng, Mengzhou Xia, and Danqi Chen.
\newblock Simpo: Simple preference optimization with a reference-free reward.
\newblock 2024.
\newblock arXiv:2405.14734.

\bibitem[Munos et~al.(2023)Munos, Valko, Calandriello, Azar, Rowland, Guo, Tang, Geist, Mesnard, and Michi]{Nashing_Learning}
R{\'e}mi Munos, Michal Valko, Daniele Calandriello, Mohammad~Gheshlaghi Azar, Mark Rowland, Zhaohan~Daniel Guo, Yunhao Tang, Matthieu Geist, Thomas Mesnard, and Andrea Michi.
\newblock Nash learning from human feedback.
\newblock 2023.
\newblock arXiv:2312.00886.

\bibitem[Ouyang et~al.(2022)Ouyang, Wu, Jiang, Almeida, Wainwright, Mishkin, Zhang, Agarwal, Slama, and Ray]{RLHF2022}
Long Ouyang, Jeffrey Wu, Xu~Jiang, Diogo Almeida, Carroll Wainwright, Pamela Mishkin, Chong Zhang, Sandhini Agarwal, Katarina Slama, and Alex Ray.
\newblock Training language models to follow instructions with human feedback.
\newblock In \emph{Neurips}, volume~35, pp.\  27730--27744, 2022.

\bibitem[Pitman \& Tang(2019)Pitman and Tang]{PT19}
Jim Pitman and Wenpin Tang.
\newblock Regenerative random permutations of integers.
\newblock \emph{Ann. Probab.}, 47\penalty0 (3):\penalty0 1378--1416, 2019.

\bibitem[Rafailov et~al.(2023)Rafailov, Sharma, Mitchell, Ermon, Manning, and Finn]{DPO}
Rafael Rafailov, Archit Sharma, Eric Mitchell, Stefano Ermon, Christopher~D Manning, and Chelsea Finn.
\newblock Direct preference optimization: Your language model is secretly a reward model.
\newblock In \emph{Neurips}, volume~36, 2023.

\bibitem[Schulman et~al.(2017)Schulman, Wolski, Dhariwal, Radford, and Klimov]{PPO}
John Schulman, Filip Wolski, Prafulla Dhariwal, Alec Radford, and Oleg Klimov.
\newblock Proximal policy optimization algorithms.
\newblock 2017.
\newblock arXiv:1707.06347.

\bibitem[Stiennon et~al.(2020)Stiennon, Ouyang, Wu, Ziegler, Lowe, Voss, Radford, Amodei, and Christiano]{LSHF2020}
Nisan Stiennon, Long Ouyang, Jeffrey Wu, Daniel Ziegler, Ryan Lowe, Chelsea Voss, Alec Radford, Dario Amodei, and Paul~F Christiano.
\newblock Learning to summarize with human feedback.
\newblock In \emph{Neurips}, volume~33, pp.\  3008--3021, 2020.

\bibitem[Tajwar et~al.(2024)Tajwar, Singh, Sharma, Rafailov, Schneider, Xie, Ermon, Finn, and Kumar]{tajwar2024preference}
Fahim Tajwar, Anikait Singh, Archit Sharma, Rafael Rafailov, Jeff Schneider, Tengyang Xie, Stefano Ermon, Chelsea Finn, and Aviral Kumar.
\newblock Preference fine-tuning of {L}{L}{M}s should leverage suboptimal, on-policy data.
\newblock 2024.
\newblock arXiv:2404.14367.

\bibitem[Tang(2019)]{tang2019mallows}
Wenpin Tang.
\newblock Mallows ranking models: maximum likelihood estimate and regeneration.
\newblock In \emph{ICML}, pp.\  6125--6134, 2019.

\bibitem[Tang et~al.(2024)Tang, Guo, Zheng, Calandriello, Munos, Rowland, Richemond, Valko, Pires, and Piot]{GPO}
Yunhao Tang, Zhaohan~Daniel Guo, Zeyu Zheng, Daniele Calandriello, R{\'e}mi Munos, Mark Rowland, Pierre~Harvey Richemond, Michal Valko, Bernardo~{\'A}vila Pires, and Bilal Piot.
\newblock Generalized preference optimization: A unified approach to offline alignment.
\newblock 2024.
\newblock arXiv:2402.05749.

\bibitem[Tunstall et~al.(2023)Tunstall, Beeching, Lambert, Rajani, Rasul, Belkada, Huang, von Werra, Fourrier, and Habib]{Zephyr}
Lewis Tunstall, Edward Beeching, Nathan Lambert, Nazneen Rajani, Kashif Rasul, Younes Belkada, Shengyi Huang, Leandro von Werra, Cl{\'e}mentine Fourrier, and Nathan Habib.
\newblock Zephyr: Direct distillation of {L}{M} alignment.
\newblock 2023.
\newblock arXiv:2310.16944.

\bibitem[Wang et~al.(2024{\natexlab{a}})Wang, Zheng, Chen, Liu, Dou, Huang, Shen, Jin, Zhou, and Shi]{Secrets_of_RLHF_II}
Binghai Wang, Rui Zheng, Lu~Chen, Yan Liu, Shihan Dou, Caishuang Huang, Wei Shen, Senjie Jin, Enyu Zhou, and Chenyu Shi.
\newblock Secrets of rlhf in large language models part {I}{I}: Reward modeling.
\newblock 2024{\natexlab{a}}.
\newblock arXiv:2401.06080.

\bibitem[Wang et~al.(2023)Wang, Jiang, Yang, Liu, and Chen]{f-DPO}
Chaoqi Wang, Yibo Jiang, Chenghao Yang, Han Liu, and Yuxin Chen.
\newblock Beyond reverse {K}{L}: Generalizing direct preference optimization with diverse divergence constraints.
\newblock 2023.
\newblock arXiv:2309.16240.

\bibitem[Wang et~al.(2024{\natexlab{b}})Wang, Xiong, Xie, Zhao, and Zhang]{wang2024interpretable}
Haoxiang Wang, Wei Xiong, Tengyang Xie, Han Zhao, and Tong Zhang.
\newblock Interpretable preferences via multi-objective reward modeling and mixture-of-experts.
\newblock 2024{\natexlab{b}}.
\newblock arXiv:2406.12845.

\bibitem[Weinshall et~al.(2018)Weinshall, Cohen, and Amir]{weinshall2018curriculum}
Daphna Weinshall, Gad Cohen, and Dan Amir.
\newblock Curriculum learning by transfer learning: Theory and experiments with deep networks.
\newblock In \emph{International conference on machine learning}, pp.\  5238--5246. PMLR, 2018.

\bibitem[Winata et~al.(2024)Winata, Zhao, Das, Tang, Yao, Zhang, and Sahu]{winata2024preference}
Genta~Indra Winata, Hanyang Zhao, Anirban Das, Wenpin Tang, David~D Yao, Shi-Xiong Zhang, and Sambit Sahu.
\newblock Preference tuning with human feedback on language, speech, and vision tasks: A survey.
\newblock 2024.
\newblock arXiv:2409.11564.

\bibitem[Xu et~al.(2024)Xu, Sharaf, Chen, Tan, Shen, Van~Durme, Murray, and Kim]{CPO}
Haoran Xu, Amr Sharaf, Yunmo Chen, Weiting Tan, Lingfeng Shen, Benjamin Van~Durme, Kenton Murray, and Young~Jin Kim.
\newblock Contrastive preference optimization: Pushing the boundaries of {L}{L}{M} performance in machine translation.
\newblock 2024.
\newblock arXiv:2401.08417.

\bibitem[Zhai et~al.(2023)Zhai, Tong, Li, Cai, Qu, Lee, and Ma]{zhai2023investigating}
Yuexiang Zhai, Shengbang Tong, Xiao Li, Mu~Cai, Qing Qu, Yong~Jae Lee, and Yi~Ma.
\newblock Investigating the catastrophic forgetting in multimodal large language models.
\newblock 2023.
\newblock arXiv:2309.10313.

\bibitem[Zhao et~al.(2023)Zhao, Joshi, Liu, Khalman, Saleh, and Liu]{zhao2023slic}
Yao Zhao, Rishabh Joshi, Tianqi Liu, Misha Khalman, Mohammad Saleh, and Peter~J Liu.
\newblock Slic-hf: Sequence likelihood calibration with human feedback.
\newblock 2023.
\newblock arXiv:2305.10425.

\bibitem[Zheng et~al.(2023)Zheng, Dou, Gao, Hua, Shen, Wang, Liu, Jin, Liu, and Zhou]{Secrets_of_RLHF_I}
Rui Zheng, Shihan Dou, Songyang Gao, Yuan Hua, Wei Shen, Binghai Wang, Yan Liu, Senjie Jin, Qin Liu, and Yuhao Zhou.
\newblock Secrets of {R}{L}{H}{F} in large language models part {I}: {P}{P}{O}.
\newblock 2023.
\newblock arXiv:2307.04964.

\bibitem[Ziegler et~al.(2019)Ziegler, Stiennon, Wu, Brown, Radford, Amodei, Christiano, and Irving]{FTLM2020}
Daniel~M Ziegler, Nisan Stiennon, Jeffrey Wu, Tom~B Brown, Alec Radford, Dario Amodei, Paul Christiano, and Geoffrey Irving.
\newblock Fine-tuning language models from human preferences.
\newblock 2019.
\newblock arXiv:1909.08593.

\end{thebibliography}
\bibliographystyle{iclr2025_conference}

\newpage
\appendix
\section*{Appendix}
\section{Prompts with low/high negative log dispersion estimate}
\label{app: more examples for prompts with low or high dispersion}
\subsection{Examples of prompts with low negative log dispersion estimate}

See Table \ref{High dispersion prompts} for the examples.

\begin{table}[!htbp]
\centering
\small
\begin{tabular}{>{\raggedright\arraybackslash}p{1.5cm}p{10.5cm}}
\toprule
Human: & I need some ideas on repurposing old magazines.\\

Assistant: & How about using a magazine as a dustcover for a book?\\

Human: & It will take some effort but it could work. What else can you think of?\\

\midrule
Human: & Why do you think western society idolizes youth so much, as opposed to asian society, which idolizes elders?\\

Assistant: & Humans throughout history have consistently held important values and traits in high esteem, including trustworthiness, compassion, intelligence, and wisdom.  Given that younger people have the potential to develop these traits while older people have already developed them, it makes sense that older people’s virtues are embraced in some cultures while younger people’s virtues are embraced in other cultures.  It’s important to remember that cultures throughout the world are complex and heterogeneous, so it’s possible for some to prioritize older people yet also to respect some of the important qualities and values held by younger people.  The value we place on older people will vary inversely with the value we place on young people, and vice versa, which helps explain why different cultures come to elevate one or the other.\\

Human: & That's a fair assessment, but I'm looking for more specific reasons. Let's refocus on just the west. Why are we so youth-focused?\\

\midrule

Human:& what are some cool ocean activities?\\

\midrule

Human:& Why do almost all cartoon characters tend to only have 4 fingers? Is it some unwritten rule in the animation industry to do that?\\

\midrule

Human:& Can you provide 3 bullet points for how to teach a child how to set the dinner table?\\

\midrule

Human:& I'd like to become a world famous writer. Any suggestions on how to get there?\\
\midrule

Human:& My friend is a University Lecturer with a Ph.D. in Business. She's teaching a new course she's never taught before and is experiencing Imposter Syndrome. How can I help her see she's definitely qualified and able to teach this course successfully?\\
\bottomrule
\end{tabular}
\vspace{0.2cm}
\caption{Prompts with low neg-log dispersion or high dispersion, i.e. human are likely to have more disagreement about the answers.}
\label{High dispersion prompts}
\end{table}

\subsection{Examples of prompts with high negative log dispersion estimate}
See Table \ref{Low dispersion prompts} for the examples.
\begin{table}[!htbp]
\centering
\begin{tabular}{>{\raggedright}p{3cm}p{12cm}}
\toprule
Human: & Have you heard of Summer Palace, a historical site in China?\\

Assistant:& Yes, Summer Palace is a well-known historical site in China.  It was originally built in the eighteenth century as a summer retreat for the royal family, and is renowned for its beautiful gardens and temples.  Summer Palace is now a popular tourist destination, attracting visitors with its historic architecture and natural beauty.  There are also many tours and exhibits at the site, offering visitors an in-depth look at its history.\\

Human: & In which city is it located?\\

\midrule
Human: & Translate the following into slang: Good morning, how are you today? I have to go to work later.\\

Assistant: & Hey, how’s it going? I’m doing pretty good today. I have to head to work later.\\

Human: & Now capitalize every word of that response.\\

\midrule

Human: & Please tell me about The Battle of Waterloo.\\

Assistant: & In 1815, Napoleon led his troops and won against the British and Prussian armies.\\

Human: & No.  Napoleon lost the battle.\\

Assistant: & OK, in 1815, Napoleon led his troops and lost against the British and Prussian armies.\\

Human: & Was the battle in Belgium?\\

\midrule

Human: & Are there any NASA space missions going on currently?\\

Assistant: & As of now, no.\\

Human: & Are you sure?\\

Assistant: & NASA doesn't have any plans for missions in the immediate future. However, there are several missions on the books that might eventually get funded.\\

Human: & Thank you for this information.\\

\bottomrule
\end{tabular}
\caption{Prompts with high neg-log dispersion or low dispersion, i.e. human are likely to have less disagreement about the answers.}
\label{Low dispersion prompts}
\end{table}

\newpage

\section{Proofs}\label{sec:proofs}

\textit{Proof of Proposition \ref{prop:Mallows-theta consistent pairwise prob}.} \\

\cite{mallows} showed that the paired-comparison probability
\begin{equation}
    \mathbb{P}\left(\mu(y_1\mid x)<\mu\left(y_2\mid x\right)\right)
= \frac{1}{1+ \exp{(-2\log \phi(x) \left(\mu_0\left(y_1\mid x\right)-\mu_0\left(y_2\mid x\right)\right))}}.
\end{equation}
yields the Mallows-$\theta$ model. 
Here we sketch a proof for ease of reference. 
Write $\gamma:=-2\log \phi(x)$ for simplicity. Note that
\begin{equation}
\label{Mallows-theta pairwise probability}
   \mathbb{P}\left(y_1 \text{ is preferred to }y_2\right)
= \frac{\exp{(- \gamma \mu_0\left(y_1\right))}}{\exp{(-\gamma \mu_0\left(y_1\right))}+\exp{(-\gamma \mu_0\left(y_2\right))}}.
\end{equation}
By \Eqref{Mallows-theta pairwise probability}, the probability of observing a (full) ranking $\mu$ is:
\begin{equation}
\begin{aligned}
\mathbb{P}(\mu) &\propto \prod_{i<j}\mathbb{P}(\mu(i)\text{ is preferred to }\mu(j))
\propto \exp \left(\gamma  \sum_{i=1}^n (n-i) \left(-\mu_0(\mu^{-1}(i))\right)\right) \\
&\propto \exp \left(\gamma  \sum_{i=1}^n i \, \mu_0(\mu^{-1}(i))\right).
\end{aligned}
\end{equation}
Moreover, we have:  
\begin{equation}
\label{eq:completesq}
2\sum_{i=1}^n i \, \mu_0(\mu^{-1}(i))=-\sum_{i=1}^n (\mu_0(\mu^{-1}(i))-i)^2 + C = -\sum_{i=1}^n (\mu(i)-\mu_0(i))^2 + C,
\end{equation}
where $C = \sum_{i = 1}^n \left(i^2 + \mu_0(\mu^{-1}(i))^2\right) = n(n+1)(2n+1)/3$ is a constant, 
and the second equality is due to the right invariance of the Spearman's rho. 
Combining \Eqref{Mallows-theta pairwise probability} and \Eqref{eq:completesq} yields:
\begin{equation}
\begin{aligned}
\mathbb{P}(\mu) \propto e^{-\frac{\gamma}{2}\sum_{i=1}^n (\mu(i)-\mu_0(i))^2} \propto \phi(x)^{\sum_{i=1}^n (\mu(i)-\mu_0(i))^2}.
\end{aligned}
\end{equation}
This is exactly the Mallows-$\theta$ model.

\textit{Proof of Theorem \ref{thm:mallowspo_theta}}\\

To obtain \eqref{Mallows DPO-theta objective}, recall that given $x$, the probability of output $y_1$ being preferred than $y_2$ is:
\begin{align}
p^*\left(y_1\succ y_2\mid x\right) &= \mathbb{P}\left(\mu(y_1\mid x)<\mu\left(y_2\mid x\right)\right) \\
&=g_{d,\phi(x)}(-\mu_0(y_1 \mid x) - (-\mu_0(y_2\mid x))) \label{eq:mallowspo_pf_eq1}
\end{align}
Therefore, the modeling of such a ranking distribution requires the conditional central ranking $\mu_0(\cdot \mid x)$. However. since finding $\mu_0(\cdot \mid x)$ is computationally hard \citep{tang2019mallows}, to tackle this challenge, we explore a different path. Recall that in RLHF, we optimize the following objective:
\begin{equation*}
\max _{\pi} \mathbb{E}_{x \sim \mathcal{D}}\left[\mathbb{E}_{y \sim \pi_\theta(y \mid x)}\left[r^*(x, y)\right]-\beta \mathrm{KL}\left(\pi(\cdot\mid x) \| \pi_{\mathrm{ref}}(\cdot \mid x)\right)\right],
\end{equation*}
where $r^*(x, y)$ is the true underlying reward. By letting $r^*(x, y) = -\mu_0(y \mid x)$, we now turn to optimize the following objective:
\begin{equation*}
\max _{\pi} \mathbb{E}_{x \sim \mathcal{D}}\left[\mathbb{E}_{y \sim \pi_\theta(y \mid x)}\left[-\mu_0(y \mid x)\right]-\beta \mathrm{KL}\left(\pi(\cdot\mid x) \| \pi_{\mathrm{ref}}(\cdot \mid x)\right)\right].
\end{equation*}
As shown in section A.1 of \cite{DPO}, the optimum of such a KL-constrained reward maximization objective has the form of
$$\pi_{\mu_0}(y \mid x) = \frac{1}{Z(x)}\pi_{\mathrm{ref}}(x) \exp\left(-\frac{\mu_0(y \mid x)}{\beta}\right),$$
where $Z(x)$ is the partition function to ensure $\pi_{\mu_0}(y \mid x)$ to be a probability distribution. 
By moving terms, we have
\begin{equation}\label{eq:mallowspo_pf_eq2}
-\mu_0(y \mid x) = \beta \log \frac{\pi_{\mu_0}(y \mid x)}{\pi_{\text {ref }}(y \mid x)}+\beta \log Z(x).  
\end{equation}
Combining (\ref{eq:mallowspo_pf_eq1}) and (\ref{eq:mallowspo_pf_eq2}) gives us 
$$p^*\left(y_1\succ y_2\mid x\right) = g_{d,\phi(x)}\left( \beta \log \frac{\pi_{\mu_0}(y_1 \mid x)}{\pi_{\text {ref }}(y_1 \mid x)} -  \beta \log \frac{\pi_{\mu_0}(y_2 \mid x)}{\pi_{\text {ref }}(y_2 \mid x)}\right).$$
To maximize the likelihood estimation, our objective becomes 
$$\min_{\pi_{\mu_0}} -\mathbb{E}_{(x, y_w, y_l)\sim \mathcal{D}} \left[\log \left(g_{d, \phi(x)}\left(\beta \log \frac{\pi_{\mu_0}(y_w \mid x)}{\pi_{\mathrm{ref}}(y_w \mid x)} - \beta \log \frac{\pi_{\mu_0}(y_l \mid x)}{\pi_{\mathrm{ref}}(y_l \mid x)}\right)\right)\right].$$

To derive the MallowsPO-$\theta$ loss objective, a direct application of above derives the result, which is:
\begin{equation}
\mathcal{L}_{\text{Mallows}-\theta}(\pi_{\theta};\pi_{\text{ref}})=-\mathbb{E}\log \sigma\left(\underbrace{-2 \log \phi(x)}_{\text {dispersion of } x}\left(\beta \log \frac{\pi\left(y_w \mid x\right)}{\pi_{\mathrm{ref}}\left(y_w \mid x\right)}-\beta \log \frac{\pi\left(y_l \mid x\right)}{\pi_{\mathrm{ref}}\left(y_l \mid x\right)}\right)\right),
\end{equation}
in which the expectation is with respect to $\left(x, y_w, y_l\right) \sim \mathcal{D}$. \hfill $\blacksquare$\\

\textit{Proof of Theorem \ref{thm:mallowspo_phi}.}  \\

For the Mallows-$\phi$ model, it was shown in \cite{mallows} (see also \cite{Bu14, MW22}):
\begin{align}
    &\hspace{4mm}\mathbb{P}\left(\mu(y_1\mid x)<\mu\left(y_2\mid x\right)\right)\\
&=
\begin{cases}
\frac{\mu_0\left(y_2\mid x\right)-\mu_0\left(y_1\mid x\right)+1}{1-\phi(x)^{\mu_0\left(y_2\mid x\right)-\mu_0\left(y_1\mid x\right)+1}}-\frac{\mu_0\left(y_2\mid x\right)-\mu_0\left(y_1\mid x\right)}{1-\phi(x)^{\mu_0\left(y_2\mid x\right)-\mu_0\left(y_1\mid x\right)}}, \quad &\mu_0\left(y_2\mid x\right)-\mu_0\left(y_1\mid x\right)>0,\\
1-\frac{\mu_0\left(y_1\mid x\right)-\mu_0\left(y_2\mid x\right)+1}{1-\phi(x)^{\mu_0\left(y_1\mid x\right)-\mu_0\left(y_2\mid x\right)+1}}-\frac{\mu_0\left(y_2\mid x\right)-\mu_0\left(y_1\mid x\right)}{1-\phi(x)^{\mu_0\left(y_1\mid x\right)-\mu_0\left(y_2\mid x\right)}}, \quad &\mu_0\left(y_2\mid x\right)-\mu_0\left(y_1\mid x\right)<0,
\end{cases}
\end{align}
A similar discuss as in the proof of Theorem \ref{thm:mallowspo_theta} derives the desired result in \eqref{Mallows DPO-phi objective}.

\hfill $\blacksquare$

\textit{Proof of Proposition \ref{thm:MallowsPO_theta_as_weigted_DPO}.}  \\

The proof follows from the derivation of the equivalence between RLHF and DPO, as now the optimal policy satisfies
\begin{equation*}
c(x)^{-1}r(x, y)=\beta \log \frac{\pi_r(y \mid x)}{\pi_{\text {ref }}(y \mid x)}+\beta \log Z(x).
\end{equation*}

\hfill $\blacksquare$

\textit{Proof of Theorem \ref{thm:MallowsPO_as_psiPO}.}  \\

(i) With the Bradley-Terry connection as mentioned above, we have
\begin{equation}
\begin{aligned}
\mathbb{E}_ {y_2 \sim \tilde{\pi}}\left[f(x)\Psi\left(p^*\left(y_1 \succ y_2\mid x\right)\right)\right] & =\mathbb{E}_ {y_2 \sim \tilde{\pi}}\left[f(x)\Psi\left(\frac{e^{r(x, y_1)}}{e^{r(x, y_1)}+e^{r\left(x, y_2\right)}}\right)\right] \\
& =\mathbb{E}_ {y_2 \sim \tilde{\pi}}\left[f(x)\left(r(x, y_1)-r\left(x, y_2\right)\right)\right] \\
& =f(x)r(x, y_1)-f(x)\mathbb{E}_ {y_2 \sim \tilde{\pi}}\left[r\left(x, y_2\right)\right] ,
\end{aligned}
\end{equation}
which is a weighted reward of DPO, up to an additive constant.
It follows that the optimal policy of the generalized $\Psi$PO \eqref{eq:generalpsi2}
is the same as that of MallowsPO-$\theta$ by Theorem \ref{thm:MallowsPO_theta_as_weigted_DPO}. The same argument also proves
(ii). 

\hfill $\blacksquare$
\newpage

\section{More Examples of Mallows Ranking Model}
\label{app:Mallows Examples}
Here we provide more detailed concrete examples and corresponding computations for Mallows Ranking Model. We also illustrate the implicit effects of the total number of items $n$. 

Recall that, as defined in Section , for a positive integer $n$ that represents e.g., $n$ possible items, let $\mathfrak{S}_n$ be the set of permutations of $[n]=\{1, \ldots, n\}$ and the space of rankings. 
The probability of observing a ranking $\mu$ (which represents the preference of $n$ items, e.g., the top ranked item is preferred over the others) under the Mallows Ranking Model is:
\begin{equation}
\mathbb{P}_{\phi, \mu_0, d}(\mu):= \phi^{d\left(\mu, \mu_0\right)}/Z(\phi, d) \quad \text { for } \mu \in \mathfrak{S}_n,
\end{equation}
where $\phi \in (0,1]$ is the dispersion parameter,
$\mu_0$ is the central ranking,
$d(\cdot, \cdot)$ is a discrepancy function that is right invariant: $d(\mu_1, \mu_2)=d\left(\mu_1 \circ \mu_2^{-1}, i d\right)$ for $\mu_1, \mu_2 \in \mathfrak{S}_n$, for which:

$\bullet$ Mallows-$\theta$ model: $d(\mu_1, \mu_2)=\sum_{i=1}^n(\mu_1(i)-\mu_2(i))^2$ is the Spearman's rho,

$\bullet$ Mallows-$\phi$ model: $d(\mu_1, \mu_2)=\operatorname{inv}\left(\mu_1 \circ \mu_2^{-1}\right)$, in which $\operatorname{inv}(\mu):=\#\left\{(i, j) \in[n]^2: i<j\right.$, and $\left.\mu(i)>\mu(j)\right\}$ is the number of inversions of $\mu$, 

and $Z(\phi, d):=\sum_{\mu \in \mathfrak{S}_n} \phi^{d\left(\mu, \mu_0\right)}$ is the normalizing constant.

\subsection{Mallows-$\theta$ model}
To both qualitatively illustrate Mallows-$\theta$ Ranking Model and quantitatively compute the corresponding probability, here we consider a three items example, thus the total number of possible rankings are 6, and we compute the discrepancy function values, normalization constant when dispersion index is 0.5, and the resulting probability as in Table \ref{Tab: Mallows-theta probs}
:
\begin{table}[ht]
\centering
\begin{tabular}{|c|c|c|c|}
\toprule
Ranking &  Discrepancy & Normalization Constant & Probability \\
\midrule
(1,2,3) &            0 & \multirow{6}{*}{1.535} &      0.651 \\ 
(1,3,2) &            2 &                        &      0.163 \\ 
(2,1,3) &            2 &                        &      0.163 \\
(2,3,1) &            6 &                        &     0.010 \\
(3,1,2) &            6 &                        &     0.010 \\
(3,2,1) &           8 &                        &   0.0025 \\
\bottomrule
\end{tabular}
\caption{Mallows-$\theta$ Model probabilities when dispersion index $\phi=0.5$.}
\label{Tab: Mallows-theta probs}
\end{table}

\begin{figure}[!htbp]
  \centering
  \begin{minipage}{0.49\linewidth}
    \centering
    \includegraphics[width=\linewidth]{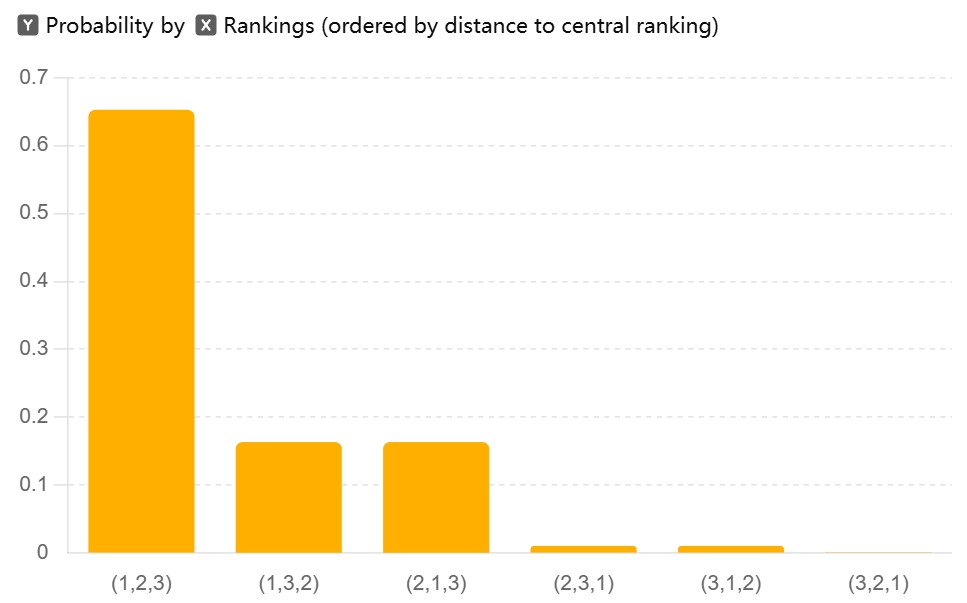}
    \subcaption{\scriptsize{Mallows-$\theta$ Probability Density function.}}
    \label{Mallows theta PDF when dispersion 0.5}
  \end{minipage} 
  \begin{minipage}{0.49\linewidth}
    \centering
    \includegraphics[width=\linewidth]{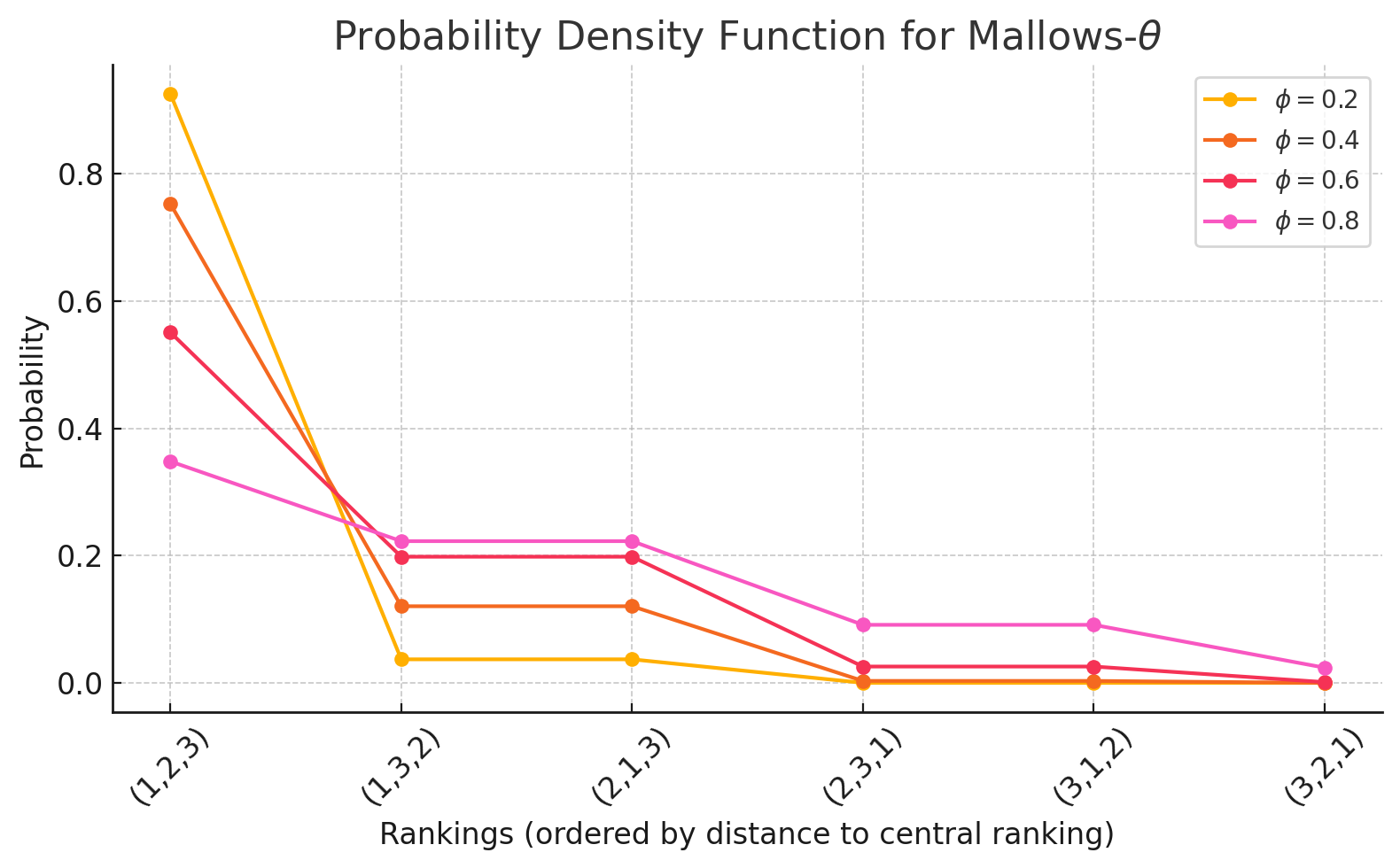}
    \subcaption{\scriptsize{PDF with respect to dispersion $\phi$}}
    \label{Mallows theta PDF vs dispersion}
  \end{minipage}
\end{figure}

We also plot the probability density function (in $x$-axis we arrange the rankings by their distance to the central ranking, the lefter the short distance to central ranking) in Figure \ref{Mallows theta PDF when dispersion 0.5}. We also plot the effects of dispersion on the resulting probabilities as in Figure \ref{Mallows theta PDF vs dispersion}. We can see as dispersion parameter goes up (close to 1), the probability density function becomes flattened.

\subsection{Mallows-$\phi$ model}
Here we consider the same example in the previous Mallows-$\theta$ except the discrepancy function is $d(\mu_1, \mu_2)=\operatorname{inv}\left(\mu_1 \circ \mu_2^{-1}\right)$. The values can be seen in Table \ref{Tab: Mallows-phi probs}.

\begin{figure}[!htbp]
  \centering
  \begin{minipage}{0.49\linewidth}
    \centering
    \includegraphics[width=\linewidth]{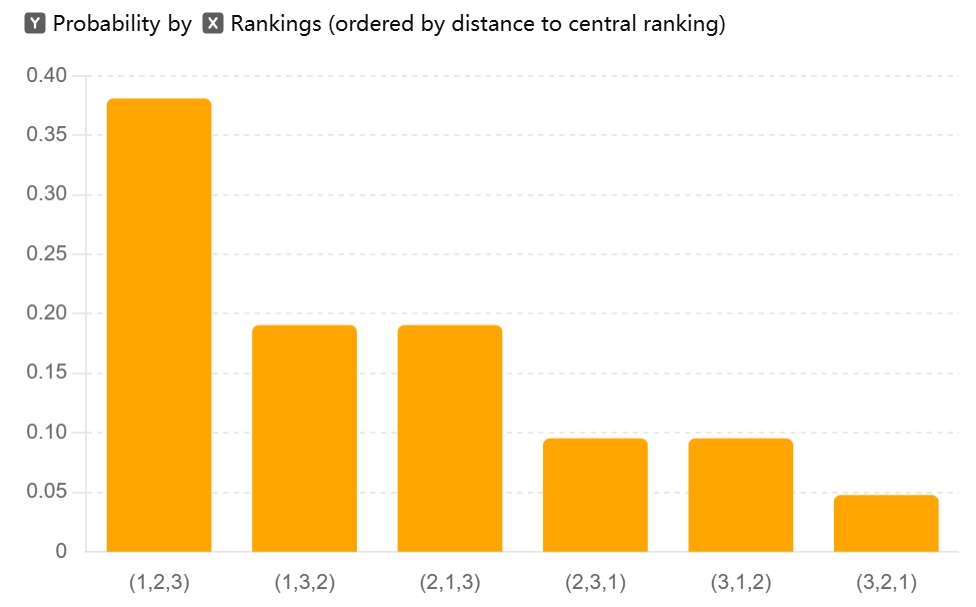}
    \subcaption{\scriptsize{Mallows-$\phi$ Probability Density function.}}
    \label{Mallows phi PDF when dispersion 0.5}
  \end{minipage} 
  \begin{minipage}{0.49\linewidth}
    \centering
    \includegraphics[width=\linewidth]{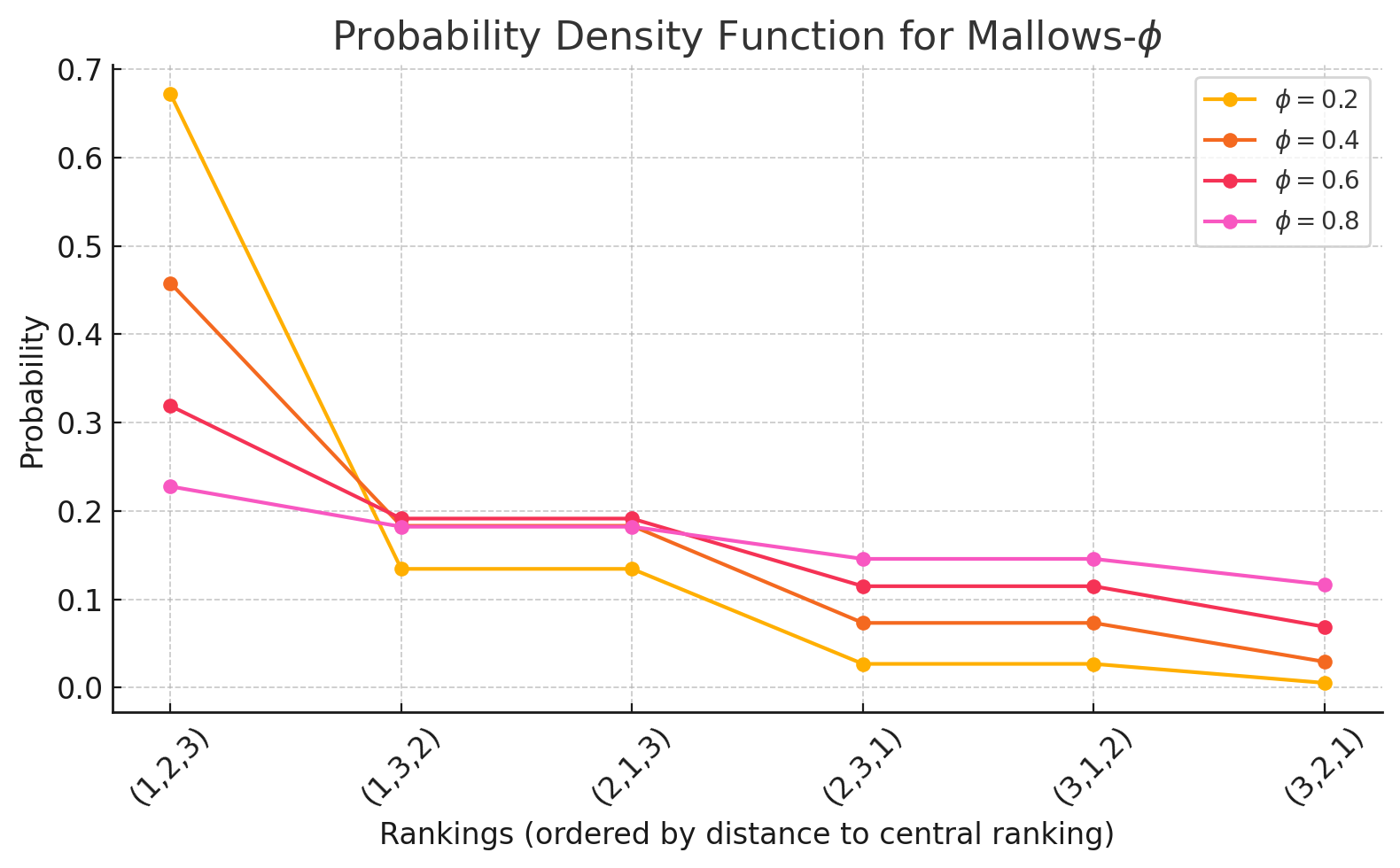}
    \subcaption{\scriptsize{PDF with respect to dispersion $\phi$}}
    \label{Mallows phi PDF vs dispersion}
  \end{minipage}
\end{figure}

\begin{table}[ht]
\centering
\begin{tabular}{|c|c|c|c|}
\toprule
Ranking &  Discrepancy & Normalization Constant & Probability \\
\midrule
(1,2,3) &            0 & \multirow{6}{*}{2.625} &      0.381 \\
(1,3,2) &            1 &                        &      0.191 \\
(2,1,3) &            1 &                        &      0.19 \\
(2,3,1) &            2 &                        &     0.095 \\
(3,1,2) &            2 &                        &     0.095 \\
(3,2,1) &            3 &                        &     0.048 \\
\bottomrule
\end{tabular}
\caption{Mallows-$\phi$ Model probabilities when dispersion index $\phi=0.5$}
\label{Tab: Mallows-phi probs}
\end{table}
Again we plot the probability density function (in $x$-axis we arrange the rankings by their distance to the central ranking, the lefter the short distance to central ranking) in Figure \ref{Mallows phi PDF when dispersion 0.5}. The effects of dispersion on the resulting probabilities as in Figure \ref{Mallows phi PDF vs dispersion}. 

\subsection{The effects of $n \gg 1$ items}

Here we also illustrate the effect of additional items for the relative probability of observing two rankings which differ only in the rankings of 3 items. 
If there are no additional items, then observing a relative order of items of values (100,1,2) with the ground truth (100,2,1) under Mallows-$\theta$ is 
(i.e., observing (1,3,2) for central ranking (1,2,3)):
$$
p(100,1,2) = Z\cdot \phi^{(3-2)^2+(2-3)^2} = Z\cdot \phi^2,
$$
which will be the same as observing (2,100,1) since:
$$
p(2,100,1) = Z\cdot \phi^{(2-1)^2+(1-2)^2} = Z\cdot \phi^2.
$$
However, if there are additional 97 items of values (100 items in total of values from 1 to 100), things will be quite different. If $\mu(i)=i$ holds for all other $i$'s from 3 to 99, we have:
$$
p(100,1,2) = Z\cdot \phi^{(99-100)^2+(100-99)^2} = Z\cdot \phi^2,
$$
while
$$
p(2,100,1) = Z\cdot \phi^{(99-1)^2+(1-99)^2} = Z\cdot \phi^{2\times 98^2}.
$$
Thus, $p(2,100,1) \ll p(100,1,2)$ because $\phi\in (0,1)$.

\newpage

\subsection{Different Mallows Models are inconsistent.}
\label{app: inconsistency between Mallows Ranking Models}

Consider two discrepancy functions used to measure the difference between two rankings (which are exactly Mallows-$\theta$ and Mallows-$\phi$ models in this paper):

A. \textbf{Squared Difference (SD):}
   \[
   \text{SD}(\mu, \mu') = \sum_{i=1}^n \left( \mu(i) - \mu'(i) \right)^2,
   \]
   where \( \mu(i) \) and \( \mu'(i) \) denote the positions of item \( i \) in the true ranking \( \mu \) and the predicted ranking \( \mu' \), respectively.

B. \textbf{Number of Inversions (NI):}
   \[
   \text{NI}(\mu, \mu') = \left| \left\{ (i, j) \mid i < j,\ \left( \mu(i) - \mu(j) \right)\left( \mu'(i) - \mu'(j) \right) < 0 \right\} \right|,
   \]
   which counts the number of pairs \( (i, j) \) where the relative order of items \( i \) and \( j \) is different in \( \mu \) and \( \mu' \). Let the ground truth ranking \( \mu \) of four items \( \{ A, B, C, D \} \) be:
\[
\mu = \left( A:1,\ B:2,\ C:3,\ D:4 \right).
\]

We consider two predicted rankings \( \mu_1 \) and \( \mu_2 \):

1. \textbf{Ranking \( \mu_1 \) and \( \mu_2 \):}
   \[
   \mu_1 = \left( A:2,\ B:1,\ C:4,\ D:3 \right),\quad
   \mu_2 = \left( A:1,\ B:4,\ C:2,\ D:3 \right).
   \]

A. \textbf{Squared Difference (SD):}

   For \( \mu_1 \):
   \[
   \begin{aligned}
   \text{SD}(\mu, \mu_1) &= (1 - 2)^2 + (2 - 1)^2 + (3 - 4)^2 + (4 - 3)^2 \\
                     &= (-1)^2 + 1^2 + (-1)^2 + 1^2 \\
                     &= 1 + 1 + 1 + 1 = 4.
   \end{aligned}
   \]

   For \( \mu_2 \):
   \[
   \begin{aligned}
   \text{SD}(\mu, \mu_2) &= (1 - 1)^2 + (2 - 4)^2 + (3 - 2)^2 + (4 - 3)^2 \\
                     &= 0^2 + (-2)^2 + 1^2 + 1^2 \\
                     &= 0 + 4 + 1 + 1 = 6.
   \end{aligned}
   \]

   \textbf{Conclusion:} \( \text{SD}(\mu, \mu_1) = 4 < \text{SD}(\mu, \mu_2) = 6 \), so \( \mu_1 \) is better than \( \mu_2 \) under the squared difference.

B. \textbf{Number of Inversions (NI):}

   The number of inversions is calculated by counting the number of pairs \( (i, j) \) where the order is reversed between \( \mu \) and \( \mu' \).

   \textbf{Pairs to Consider:}
   \[
   \{ (A,B),\ (A,C),\ (A,D),\ (B,C),\ (B,D),\ (C,D) \}.
   \]

   For \( \mu_1 \), \textbf{Inversion Count:}\\
     - \( (A,B) \): Inverted (\( \mu \): \( A < B \), \( \mu_1 \): \( A > B \)) \( \Rightarrow \) 1 inversion.\\
     - \( (C,D) \): Inverted (\( \mu \): \( C < D \), \( \mu_1 \): \( C > D \)) \( \Rightarrow \) 1 inversion.\\
     - Other pairs maintain the order.

     \textbf{Total Inversions in \( \mu_1 \):} 2 inversions.

   For \( \mu_2 \), \textbf{Inversion Count:}\\
     - \( (B,C) \): Inverted (\( \mu \): \( B < C \), \( \mu_2 \): \( B > C \)) \( \Rightarrow \) 1 inversion.\\
     - \( (B,D) \): Inverted (\( \mu \): \( B < D \), \( \mu_2 \): \( B > D \)) \( \Rightarrow \) 1 inversion.\\
     - Other pairs maintain the order.

     \textbf{Total Inversions in \( \mu_2 \):} 2 inversions.

   \textbf{Conclusion:} \( \text{NI}(\mu, \mu_1) = \text{NI}(\mu, \mu_2) = 2 \), so \( \mu_1 \) and \( \mu_2 \) are equally good under the number of inversions measure.

This example demonstrates an inconsistency between the squared difference and the number of inversions discrepancy functions:

- Under the squared difference measure, \( \mu_1 \) is better than \( \mu_2 \).\\
- Under the number of inversions measure, \( \mu_1 \) and \( \mu_2 \) are equally good.

\subsection{Unify Mallows-$\theta$ and Mallows-$\phi$ for computation}
\label{Mallows unified}
Note that the link function $g_{\phi,\phi(x)}$ in MallowsPO-$\phi$ is not continuous (or smooth) at $x = 0$, with
\vspace{2mm}
\begin{equation}
\label{Mallows Link Original two case derivative}
g^{\prime}_{\phi, \phi(x)}(s)=
\begin{cases}
\frac{1}{1 - \phi(x)^{s+1}} + \frac{(s+1) \phi^{s+1} \log\phi(x)}{(1 - \phi(x)^{s+1})^2} - \frac{1}{1 - \phi(x)^s} - \frac{s \phi(x)^s \log \phi(x)}{(1 - \phi(x)^s)^2}, \quad &s>0,\\
\frac{1}{1 - \phi(x)^{1-s}} + \frac{(1-s) \phi(x)^{1-s} \log \phi(x)}{(1 - \phi(x)^{1-s})^2} - \frac{1}{1 - \phi(x)^{-s}} + \frac{s \phi(x)^{-s} \log \phi(x)}{(1 - \phi(x)^{-s})^2}, \quad &s<0.
\end{cases}
\end{equation}
For computational purposes, we propose two smooth approximations to $g_{\phi,\phi(x)}$. 

\begin{figure}[!htb]
   \begin{minipage}{0.33\textwidth}
     \centering
       \includegraphics[width=0.8\linewidth]{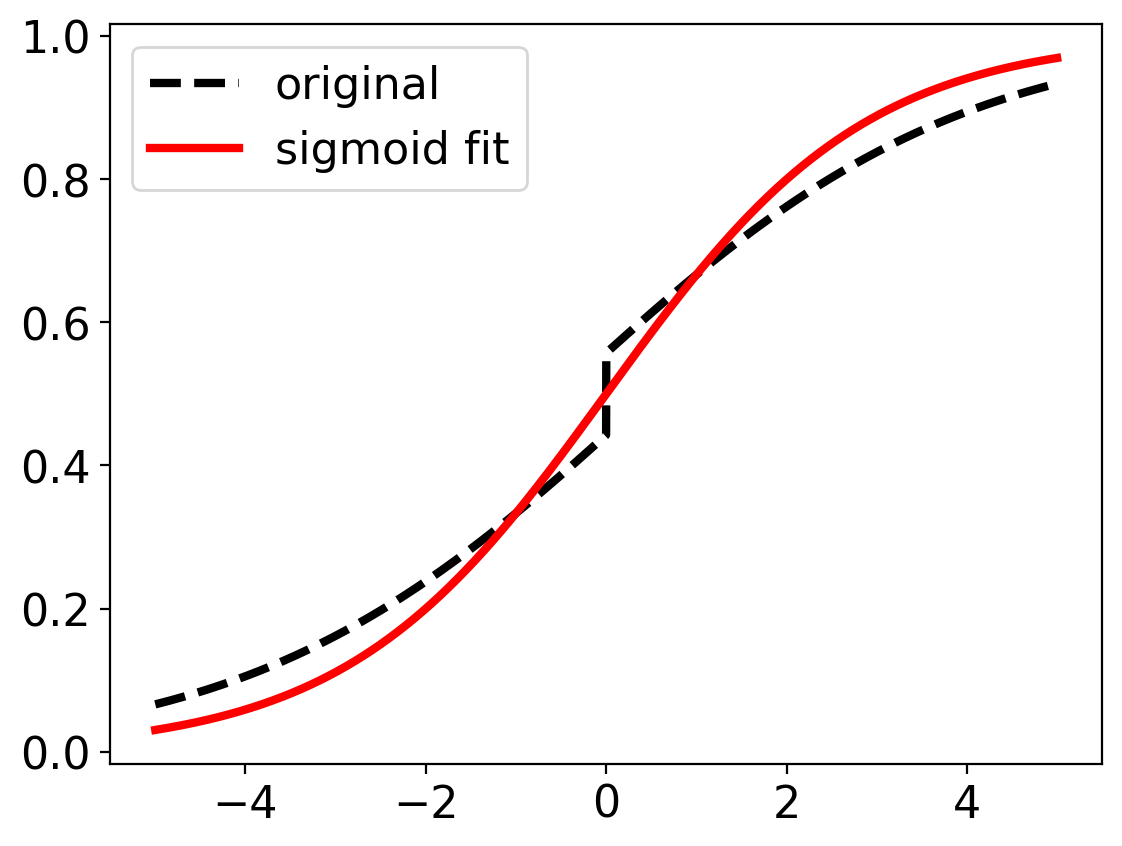}
  \caption{Sigmoid approximation}
  \label{fig: sigmoid approximation}
   \end{minipage}\hfill
    \begin{minipage}{0.33\textwidth}
    \centering
     \includegraphics[width=0.8\linewidth]{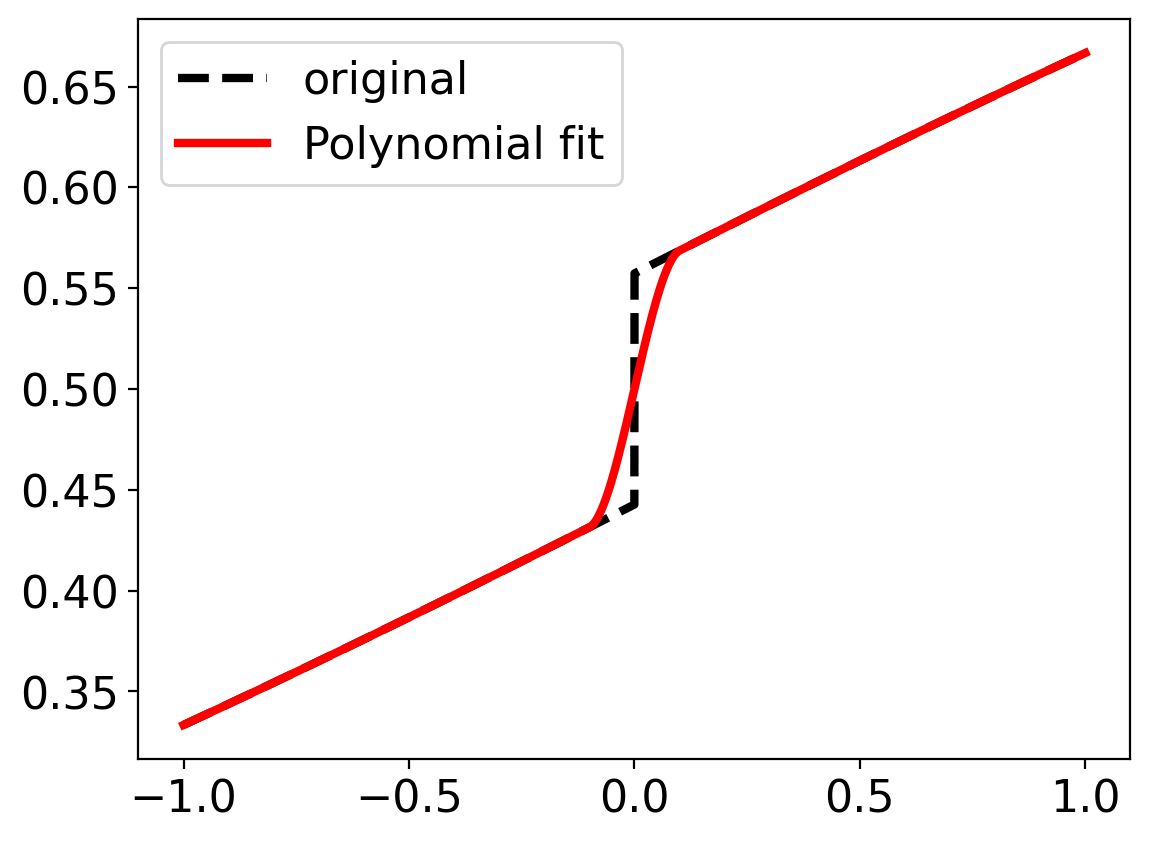}
  \caption{Poly-fitting on $\pm \epsilon$}
  \label{fig: poly_epsilon_approx}
   \end{minipage} \hfill
   \begin{minipage}{0.33\textwidth}
     \centering
       \includegraphics[width=0.8\linewidth]{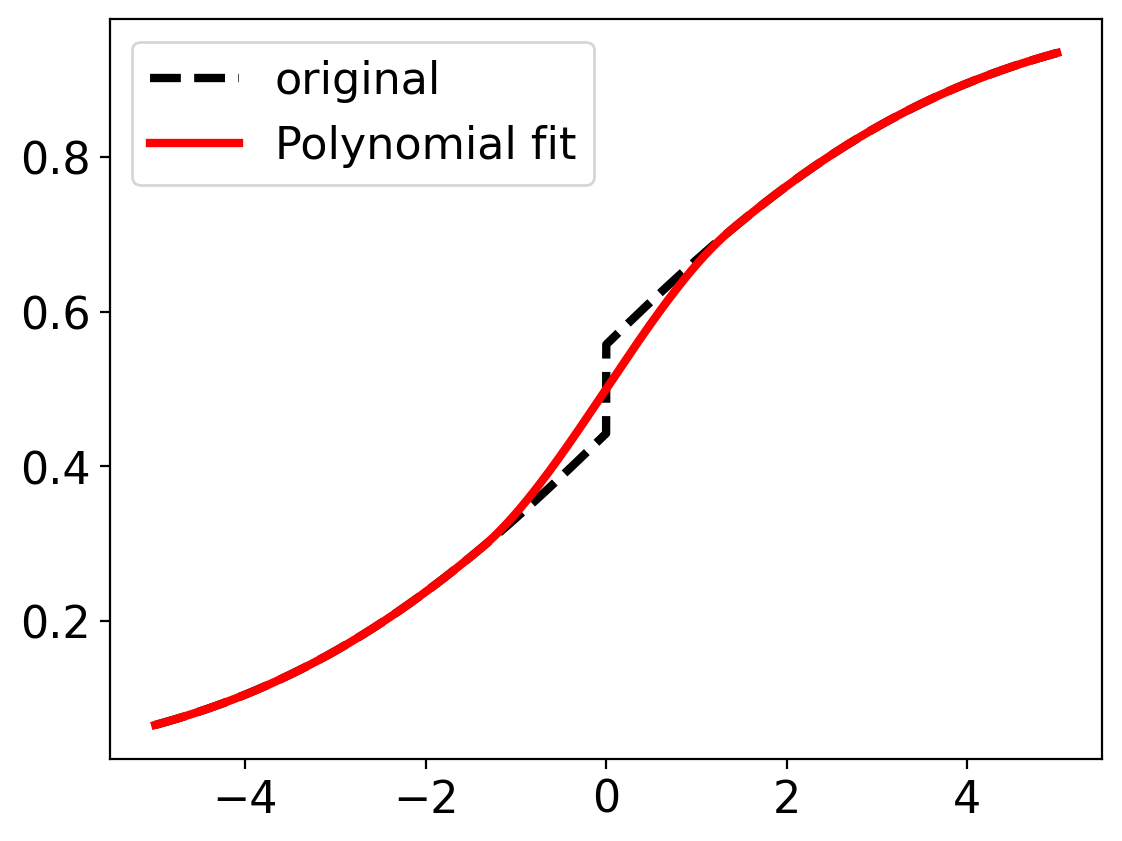}
  \caption{Poly-fitting on $ \pm 2 \log \phi$}
  \label{fig: poly_2log_phi}
   \end{minipage}
\end{figure}

(i) {\em Sigmoid approximation}: Since
$g_{\phi,\phi(x)}(1)=\frac{1}{1+\phi(x)}$,
we approximate $g_{\phi,\phi(x)}(s)$ by $\sigma_x(s):=\sigma(- s \log \phi(x))$
so that $\sigma_x(1) = g_{\phi,\phi(x)}(1)$. 
See Figure \ref{fig: sigmoid approximation} for an illustration of this approximation. 
With this approximation, 
MallowsPO-$\phi$ and MallowsPO-$\theta$ yield the same objective 
with different $\beta$'s (up to a factor of $2$).
Thus, MallowsPO-$\theta$ is just
MallowsPO-$\phi$ with sigmoid approximation.

(ii) {\em Polynomial fitting}: 
We use a polynomial of form $P(x)=a_3x^3+a_1x+a_0$ to approximate $g_{\phi,\phi(x)}$ on $[-\epsilon,\epsilon]$, with $\epsilon$ being a hyperparameter. 
We choose $\epsilon$ to be either fixed, e.g., $\epsilon = 0.1$; 
or $\epsilon = - 2 \log \phi(x)$ (e.g., $\epsilon \approx 1.4$ for $\phi(x)= 0.5$).
See Figures \ref{fig: poly_epsilon_approx}--\ref{fig: poly_2log_phi} for an illustration. 

\newpage
\section{Further Discussion on MallowsPO and DPO \citep{DPO}}

{\bf \em MallowsPO-$\theta$.} When the responses $y_1, y_2$ for a given prompt $x$ follow the preference distribution in (\ref{eq:mallows_theta_prob}), Theorem \ref{thm:mallowspo_theta} establishes the optimization objective as follows:
\begin{equation}
\begin{aligned}
& \mathcal{L}_{\mathrm{MallowsPO-\theta}}  \left(\pi ; \pi_{\mathrm{ref}}\right):= \\
& \qquad -\mathbb{E}_{\left(x, y_w, y_l\right) \sim \mathcal{D}}\left[\log \sigma \left(\underbrace{-2\log\phi(x)}_{\text{neg-log dispersion of }x}\left(\beta \log \frac{\pi\left(y_w \mid x\right)}{\pi_{\text {ref }}\left(y_w \mid x\right)}-\beta \log \frac{\pi\left(y_l \mid x\right)}{\pi_{\text {ref }}\left(y_l \mid x\right)}\right)\right)\right],
\end{aligned}
\end{equation}
where $-2\log\phi(x) \in (0, \infty)$.
To compare with the objective of DPO, where the underlying ranking model is Bradley-Terry: 
\begin{equation*}
    \mathcal{L}_{\mathrm{DPO}}\left(\pi ; \pi_{\mathrm{ref}}\right)
    :=-\mathbb{E}_{\left(x, y_w, y_l\right) \sim \mathcal{D}}\left[\log( \sigma\left(\beta \log \frac{\pi \left(y_w \mid x\right)}{\pi_{\text {ref }}\left(y_w \mid x\right)}-\beta \log \frac{\pi \left(y_l \mid x\right)}{\pi_{\text {ref }}\left(y_l \mid x\right)}\right)\right],
\end{equation*}
the objective of
MallowsPO-$\theta$ in \eqref{Mallows DPO-theta objective} 
has an extra term $-2\log \phi(x)$,
which reflects the dispersion of the prompt $x$. The similarity of MallowsPO-$\theta$ and DPO is resulted from the similarity of the foundational ranking models between Bradley-Terry and Mallows-$\theta$.

Alternatively, MallowsPO-$\theta$ can be viewd as DPO with an added 
element of prompt dispersion or disparity. Rather than interpreting MallowsPO-$\theta$ solely as $g_{d, \phi(x)} = g_{\theta, \phi(x)}$, we can reinterpret that such an objective still maintains the link function of sigmoid $\sigma(\cdot)$, but the logit difference term $\beta \log \frac{\pi \left(y_w \mid x\right)}{\pi_{\text {ref }}\left(y_w \mid x\right)}-\beta \log \frac{\pi \left(y_l \mid x\right)}{\pi_{\text {ref }}\left(y_l \mid x\right)}$ is sacled by the dispersion $-2\log\phi(x)$. This scaling introduces a nuanced modification to the original DPO framework, accounting for the influence of prompt-specific dispersion.

{\bf \em MallowsPO-$\phi$.}
Similarly, when the responses $y_1, y_2$ for a given prompt $x$ follow the preference distribution in (\ref{eq:mallows_phi_prob}), Theorem \ref{thm:mallowspo_phi} yields the following optimization objective:
\begin{equation}
\begin{aligned}
& \mathcal{L}_{\mathrm{MallowsPO-\phi}}  \left(\pi ; \pi_{\mathrm{ref}}\right):= \\
& \qquad -\mathbb{E}_{\left(x, y_w, y_l\right) \sim \mathcal{D}}\left[\log g_{\phi, \phi(x)} \left(\beta \log \frac{\pi\left(y_w \mid x\right)}{\pi_{\text {ref }}\left(y_w \mid x\right)}-\beta \log \frac{\pi\left(y_l \mid x\right)}{\pi_{\text {ref }}\left(y_l \mid x\right)}\right)\right],
\end{aligned}
\end{equation}
where $g_{\phi, \phi(x)}(s) := \frac{1-\operatorname{sgn}(s)}{2}+\operatorname{sgn}(s)\left(
\frac{|s|+1}{1-\phi(x)^{|s|+1}}-\frac{|s|}{1-\phi(x)^{|s|}}\right)$

In comparison with the DPO in \eqref{DPO objective}, 
MallowsPO-$\phi$ replaces the sigmoid function $\sigma$ with a (different) link function $g_{\phi, \phi(x)}$ that also contains the dispersion index $\phi(x)$.  

An overview of the comparison of these methods is provided in Table \ref{tab:Mallows-DPO models compared to DPO}.

\begin{table}[!htbp]
\centering
\vspace{0.2cm}
\renewcommand{\arraystretch}{1.1} %
\begin{tabular}{@{}llcc@{}}
\toprule
Model Name & Ranking model class & Dispersion & Link function \\
\midrule
DPO \citep{DPO} & Bradley-Terry \citep{BT} &  \makecell{\xmark}  & sigmoid $\sigma (\cdot)$  \\
\hline 
MallowsPO-$\theta$ & Mallows-$\theta$ \citep{mallows} & \cmark & sigmoid $\sigma (\cdot)$ \\
MallowsPO-$\phi$ & Mallows-$\phi$ \citep{mallows} &\cmark & $g_{\phi, \phi(x)}(\cdot)$ in \eqref{eq:mallows_phi_prob} \\
\bottomrule
\end{tabular}
\caption{Key Features of DPO and MallowsPO}%
\label{tab:Mallows-DPO models compared to DPO}
\end{table}
\newpage

\section{Experimental Details}

\subsection{Bandit Experiment}

In the bandit experiment detailed in Section \ref{sec:bandit}, we conduct two sub-experiments to compute the efficient frontiers using Mallow-$\phi$-DPO and DPO. The first sub-experiment varies the parameter $\beta$ while the second varies the epochs, with $\beta$'s to be a fixed set. For the first sub-experiment, we run each algorithm on a range of $\beta$ values required to compute the full efficient frontier, and for each $\beta$, we record the reward and $\text{KL}(\pi || \pi_{\text{ref}})$ of the average policy over the last 30 epochs to stabilize the results. As for the second sub-experiment, similar to the setup in \cite{DPO} and \cite{f-DPO}, we execute an ensemble of training configurations for both MallowsPO and DPO, by adopting a range of different $\beta \in \{0.05, 0.1, 0.5, 1.0\}$, and record the average reward and average $\text{KL}(\pi || \pi_{\text{ref}})$ among the four policies for every $100$ training steps. Given that we know the real reward distribution, all these quantities can be computed analytically.

\quad In terms of the training details, we use all 16 data in a single batch and adopts \texttt{SGD} as the optimizer, with learning rate of \texttt{5e-3}. To ensure convergence, we run the optimization for a large number of epochs, set to \texttt{500,000}. For MallowsPO-$\phi$, we set $\phi$ to be \texttt{0.05}.

\begin{table}[ht]
    \centering
    \caption{Reward distributions of the five arms.}
    \vspace{0.2cm}
    \begin{tabular}{|cl|cl|cl|cl|cl|}
    \hline
    \multicolumn{2}{|c|}{Arm 1}&\multicolumn{2}{c|}{Arm 2}&\multicolumn{2}{c|}{Arm 3}&\multicolumn{2}{c|}{Arm 4}&\multicolumn{2}{c|}{Arm 5}\\
    \hline
    \hline
  Reward&    Prob.& Reward& Prob.& Reward& Prob.& Reward& Prob.& Reward&Prob.\\
    \hline
          20&   0.5&30&   0.5&18&   0.5&15&  0.99& 1&0.99\\
          11&   0.5&3&   0.5&15&   0.5&10&  0.01& 4&0.01\\
    \hline
    \end{tabular}
    \label{tab: reward_dist}
\end{table}

\begin{table}[ht]
    \centering
    \caption{16 pairs of sampled preference data.}
    \vspace{0.2cm}
    \begin{tabular}{|c|c|c|c|c|c|c|c|c|c|l|l|l|l|l|l|l|}
    \hline
         Win&  3&  2&  2&  1&  3&  1&  1&  1&  4& 2& 2& 2& 1& 3& 3&4\\
         Lose&  5&  5&  5&  2&  5&  5&  4&  5&  5& 4& 1& 5& 3& 5& 4&2\\
    \hline
    \end{tabular}
    \label{tab: syn_data}
\end{table}

\subsection{Controllable Generation Experiment Details}
We follow the training setup in \cite{DPO}, and first fine-tune \texttt{GPT-2-large} on the training split of IMDB datasets until convergence to get the SFT model. The next step is different from \cite{DPO} in that we directly utilize the (offline) preference dataset from \cite{f-DPO} instead of generating pairwise preferences from the trained SFT model, as in DPO. The rest is the same: we use the pairwise preference data to fine-tune the SFT model by either DPO or MallowsPO. The evaluation metric: accuracy is obtained from a prior sentiment classifier as the ground truth reward. By default, we use \texttt{RMSprop} optimizer with a learning rate of \texttt{1e-6}, with a linear learning rate warmup from 0 to \texttt{1e-6} over the first \texttt{150} steps. The training batch size is \texttt{64}.

\subsection{Language Modeling Experiment Details}

We follow the training setup in \cite{DPO}. By default, we use \texttt{RMSprop} optimizer with a learning rate of \texttt{1e-6}, with a linear learning rate warmup from 0 to \texttt{1e-6} over the first \texttt{150} steps. The training batch size is \texttt{32}.

\subsubsection{GPT-4 Judgement Prompt}
Response quality evaluation is completed by GPT-4. The prompt for instructing GPT-4 to evaluate which response is better is particularly important. 
Thus, we use the \texttt{fastchat} package for GPT-4 evaluation, and we used their well-written \texttt{pair-v2} judge prompt. The prompt is shown as follows:

\texttt{Please act as an impartial judge and evaluate the quality of the responses provided by two AI assistants to the user question displayed below. You should choose the assistant that follows the user's instructions and answers the user's question better. Your evaluation should consider factors such as the helpfulness, relevance, accuracy, depth, creativity, and level of detail of their responses. Begin your evaluation by comparing the two responses and provide a short explanation. Avoid any position biases and ensure that the order in which the responses were presented does not influence your decision. Do not allow the length of the responses to influence your evaluation. Do not favor certain names of the assistants. Be as objective as possible. After providing your explanation, output your final verdict by strictly following this format: \textbackslash`` [[A]]\textbackslash'' if assistant A is better, \textbackslash``[[B]]\textbackslash'' if assistant B is better, and \textbackslash``[[C]]\textbackslash'' for a tie."}

\quad To ensure fairness and unbiasedness, for each pairwise input $(x, y_1, y_2)$, \texttt{fastchat} conducts two evaluation: first comparing $(y_1, y_2)$ and then comparing $(y_2, y_1)$. $y_1$ wins if and only if it wins both comparisons, or wins one comparison while the other is tied.
We compute win rate as follows:
\begin{equation*}
    \text{Win rate (Model A)} = \frac{\text{Number of samples where Model A wins}}{\text{Total number of test samples}} + 0.5 \times \frac{\text{Number of tied samples}}{\text{Total number of test samples}}
\end{equation*}

\section{Additional Results}

\subsection{ArmoRM reward model}
\label{append:Armo}
\begin{figure}[!htbp]
  \centering
  \includegraphics[width=0.9\linewidth]{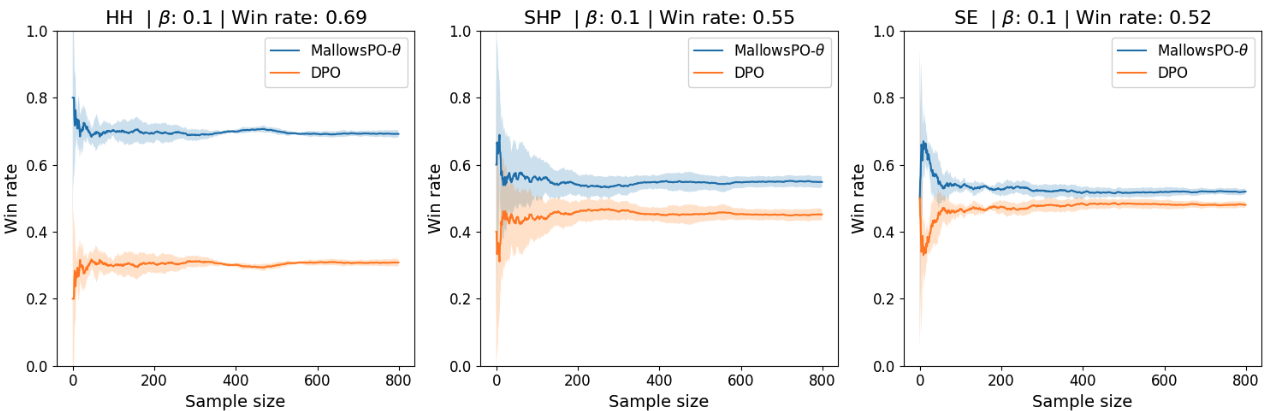}
  \caption{\small{Win rates computed by ArmoRM for responses on both the in- and out-of distribution dataset. Experiments are repeated over 5 runs.}}
  \label{fig: armoRM_results}
\end{figure}

ArmoRM \citep{wang2024interpretable} is a reward model for language modeling that utilizes multi-objective reward modeling and Mixture-of-Experts (MoE) techniques. Building on this, we also compare MallowsPO-$\theta$ with DPO in fine-tuning the Pythia-2.8B model on Anthropic-HH dataset,  setting $\beta=0.1$, as per the setting in \cite{DPO}, with ArmoRM serving as the evaluator. For evaluation, following the procedure in Section \ref{sec:id_and_ood_results},  we assess the models on the Anthropic-HH test set for in-distribution performance and on the SHP and SE datasets for out-of-distribution performance. The result indicates that model fine-tuned with MallowsPO-$\theta$ achieves consistently higher win rates than DPO across all cases, with an impressive win rate of around 70\% in the in-distribution test.

\subsection{Ablation on $\beta$ and $lr$ for MallowsPO and DPO}\label{sec:ablation}
Including the setting in Section \ref{exp:llama3}, we compare the performance of DPO and MallowsPO-$\theta$ in 6 configs by combining commonly used $\beta \in \{0.01, 0.05, 0.1\}$ and $lr \in \{e^{-6}, 5e^{-7}\}$. We find that in \textbf{5 out of 6} configs, MallowsPO-$\theta$ achieves better Length Controlled Win Rate and Win Rate.

\begin{table}[htbp!]
\centering
\small
\label{tab:more_win_rates}
\begin{tabular}{cc : cc : cc}
\toprule
\multirow{2}{*}{$\beta$} & \multirow{2}{*}{$lr$} & \multicolumn{2}{c:}{LC Win Rate} & \multicolumn{2}{c}{Win Rate} \\
\cline{3-6}
& & DPO & MallowsPO-$\theta$ & DPO & MallowsPO-$\theta$ \\
\midrule
0.1  & $e^{-6}$   & 37.14\% & 37.37\% & 35.77\% & 35.83\% \\
0.05 & $e^{-6}$   & 40.08\% & 41.08\% & 40.31\% & 41.02\% \\
0.01 & $e^{-6}$   & 27.42\% & 29.65\% & 27.03\% & 28.92\% \\
0.1  & $5e^{-7}$  & 36.09\% & 37.66\% & 35.72\% & 37.05\% \\
0.05 & $5e^{-7}$  & 40.56\% & 39.75\% & 39.89\% & 39.49\% \\
0.01 & $5e^{-7}$  & 42.55\% & \textbf{43.10\%} & 42.02\% & \textbf{43.02\%} \\
\bottomrule
\end{tabular}
\caption{Win rate comparison between DPO and MallowsPO-$\theta$ with different $\beta$ and $lr$.}
\end{table}

\subsection{Qualitative Examples}

\begin{figure}[!htbp]
  \centering
  \caption{\small{Sample responses to a prompt from the Anthropic-HH test set. GPT-4 chooses MallowsPO-$\theta$ over DPO. Both MallowsPO-$\theta$ and DPO are trained for $\beta = 0.1$.}}
  \includegraphics[width=\linewidth]{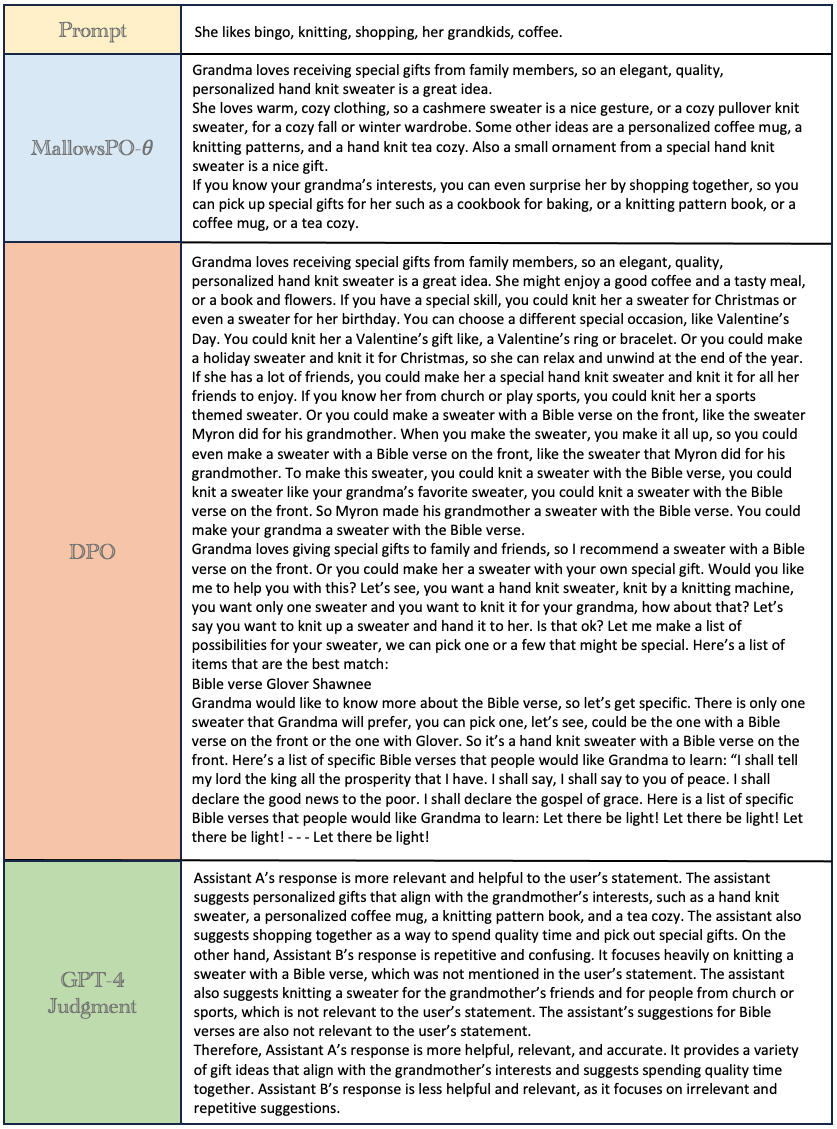}
  \label{tab:hh_1_Mallows_theta_DPO_vs_BT_DPO}
\end{figure}

\begin{figure}[!htbp]
  \centering
  \caption{\small{Sample responses to a prompt from the Anthropic-HH test set. GPT-4 chooses MallowsPO-$\phi$ over DPO. Both MallowsPO-$\phi$ and DPO are trained for $\beta = 0.1$.}}
  \includegraphics[width=\linewidth]{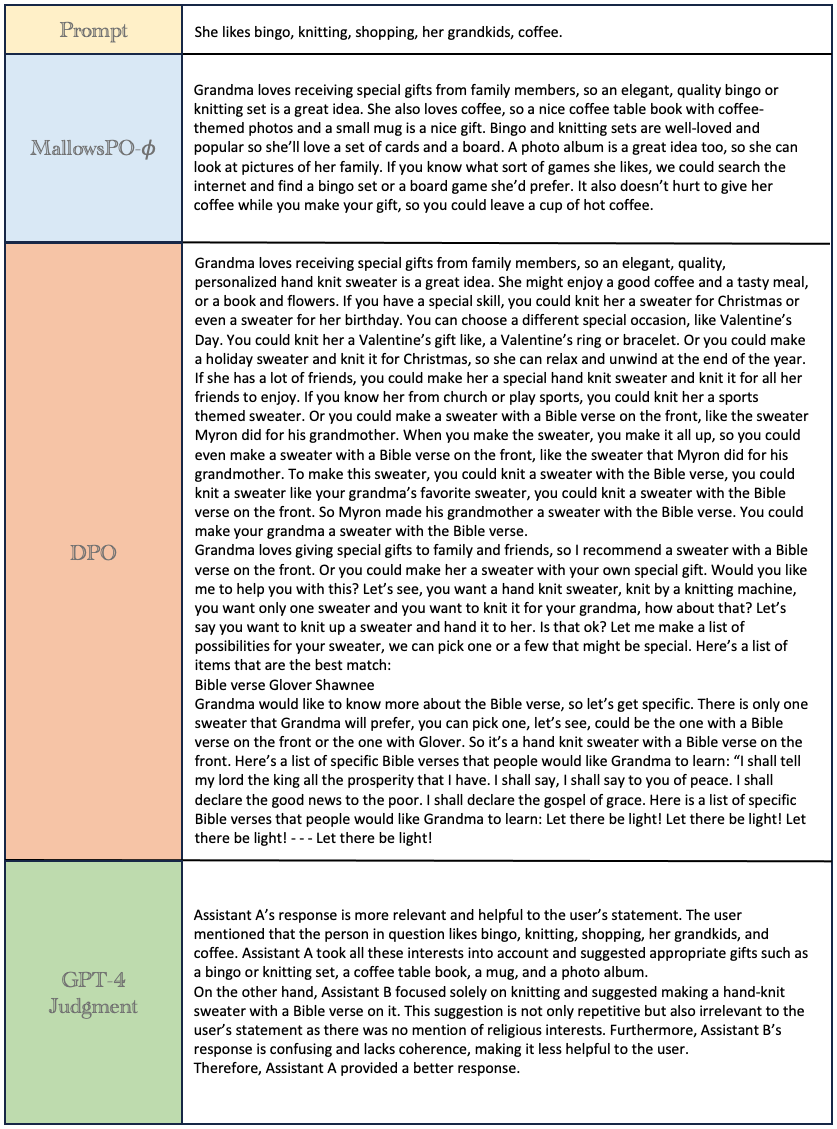}
 \label{tab:hh_1_Mallows_phi_DPO_vs_BT_DPO}
\end{figure}

\begin{figure}[!htbp]
  \centering
  \caption{\small{Sample responses to a prompt from the test set of the Stanford Human Preferences Dataset. GPT-4 chooses MallowsPO-$\theta$ over DPO. Both MallowsPO-$\theta$ and DPO are trained for $\beta = 0.1$.}}
  \includegraphics[width=\linewidth]{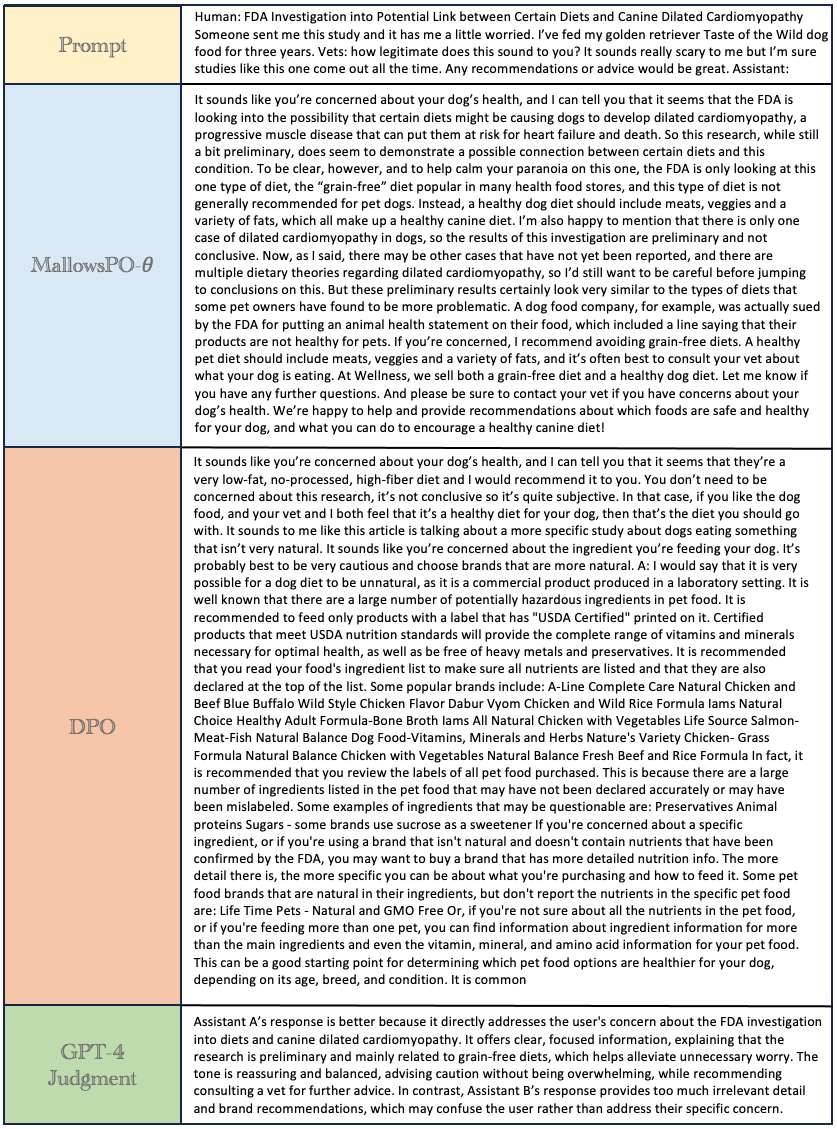}
 \label{tab:shp_1_Mallows_theta_DPO_vs_BT_DPO}
\end{figure}

\begin{figure}[!htbp]
  \centering
  \caption{\small{Sample responses to a prompt from the test set of the Stanford Human Preferences Dataset. GPT-4 chooses MallowsPO-$\theta$ over DPO. Both MallowsPO-$\theta$ and DPO are trained for $\beta = 0.5$.}}
  \includegraphics[width=\linewidth]{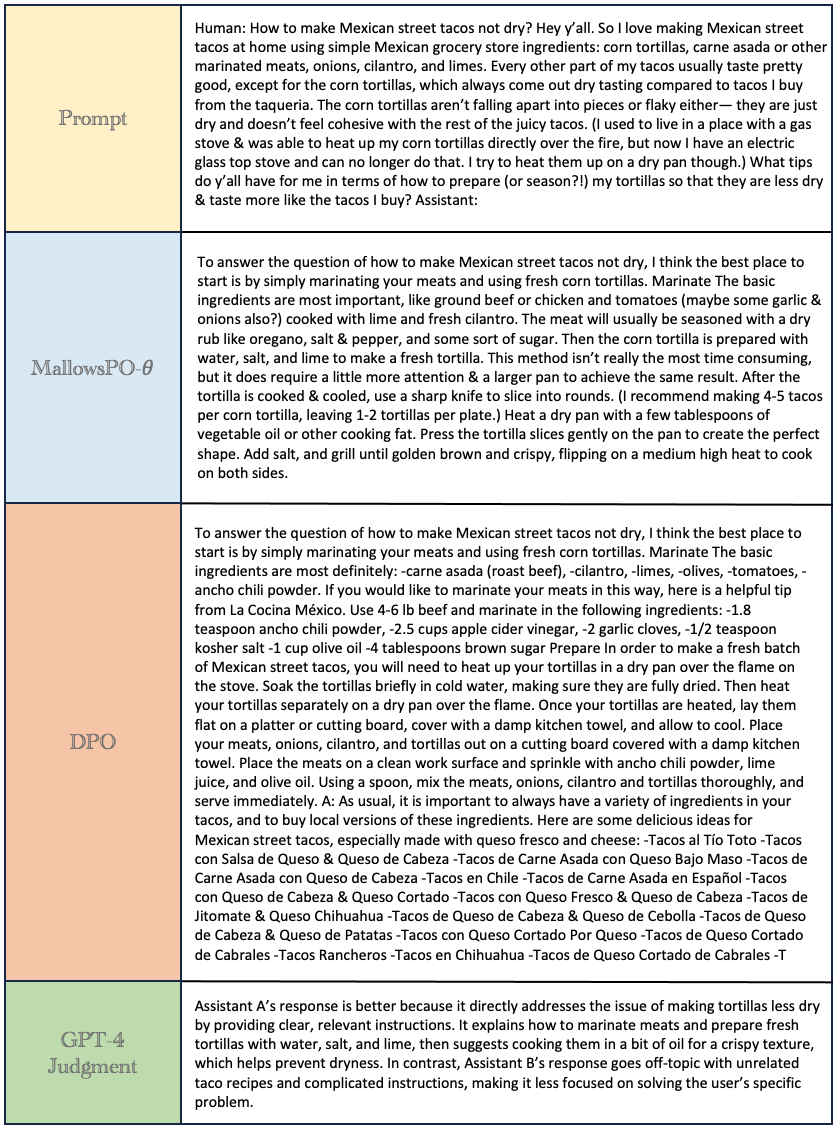}
 \label{tab:shp_5_Mallows_theta_DPO_vs_BT_DPO}
\end{figure}

\begin{figure}[!htbp]
  \centering
  \caption{\small{Sample responses to a prompt from the test set of the H4 Stack Exchange Preferences Dataset. GPT-4 chooses MallowsPO-$\theta$ over DPO. Both MallowsPO-$\theta$ and DPO are trained for $\beta = 0.1$ on Anthropic-HH dataset.}}
  \includegraphics[width=\linewidth]{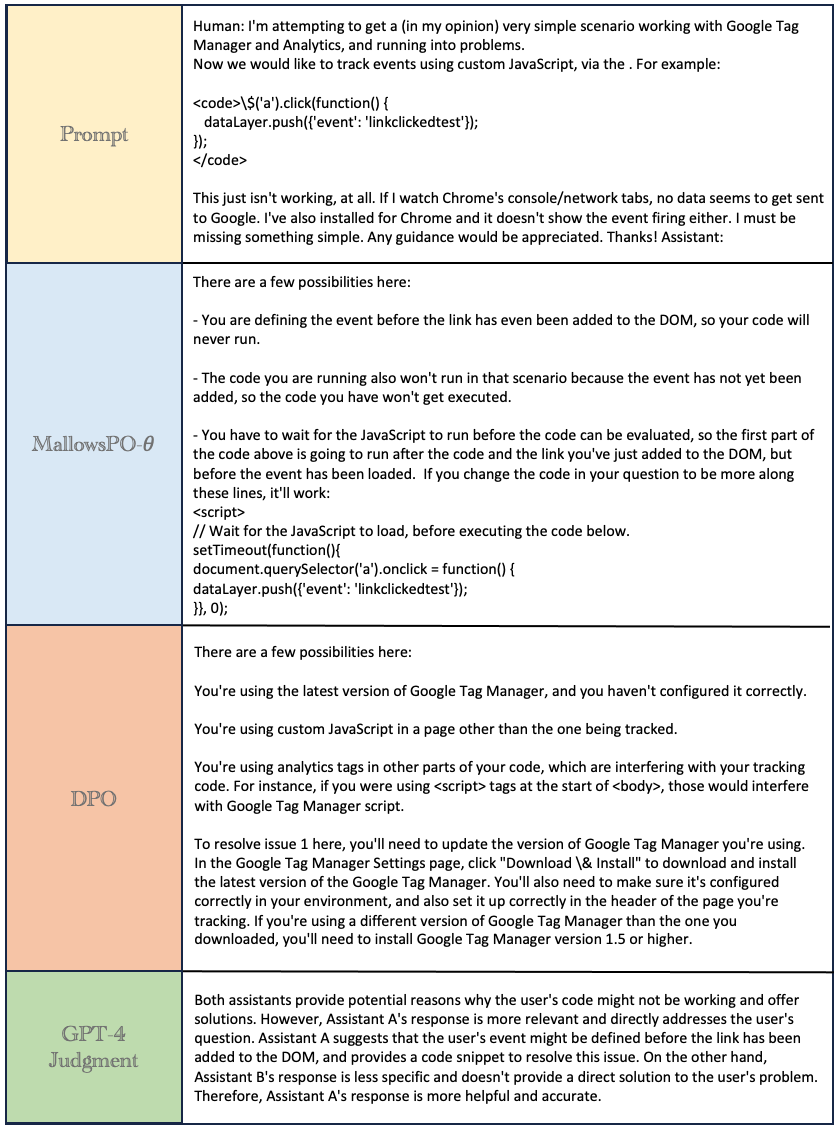}
 \label{tab:se_1_Mallows_theta_DPO_vs_BT_DPO}
\end{figure}

\begin{figure}[!htbp]
  \centering
  \caption{\small{Sample responses to a prompt from the test set of the H4 Stack Exchange Preferences Dataset. GPT-4 chooses MallowsPO-$\phi$ over DPO. Both MallowsPO-$\phi$ and DPO are trained for $\beta = 0.5$ on Anthropic-HH dataset.}}
  \includegraphics[width=\linewidth]{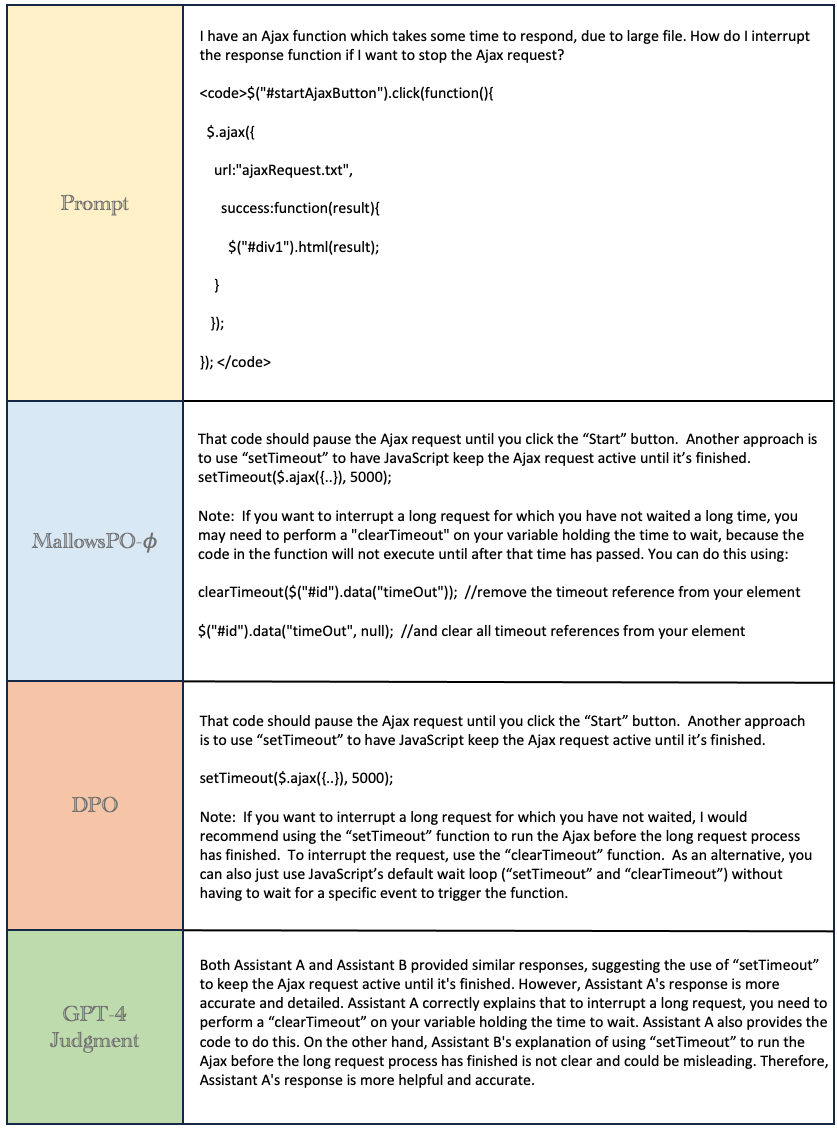}
 \label{tab:se_5_Mallows_phi_DPO_vs_BT_DPO}
\end{figure}

In this section, we present a series of examples for direct comparisons between MallowsPO variants and DPO, as shown in Tables \ref{tab:hh_1_Mallows_theta_DPO_vs_BT_DPO}--\ref{tab:se_5_Mallows_phi_DPO_vs_BT_DPO}. These tables showcase the qualitative examples of model responses both in-distribution inputs from the Anthropic-HH test set and out-of-distribution inputs from the SHP test set and the SE dataset respectively. SHP dataset covers questions/instructions in 18 different subject areas, from cooking to legal advice. SE dataset contains questions and answers from the Stack Overflow Data Dump.

To interpret, Tables \ref{tab:hh_1_Mallows_theta_DPO_vs_BT_DPO}--\ref{tab:hh_1_Mallows_phi_DPO_vs_BT_DPO} highlight that models fine-tuned with MallowsPO-$\theta$ and MallowsPO-$\phi$ demonstrate a strong grasp of contextual understanding and produce concise suggestions; however, the DPO fine-tuned model has the issue of generating repetitive and confusion response. 

The examples in Tables \ref{tab:shp_1_Mallows_theta_DPO_vs_BT_DPO}--\ref{tab:shp_5_Mallows_theta_DPO_vs_BT_DPO} show these model's responses to questions from daily life, testing their abilities in tutoring and knowledge retrieval. These results indicate that MallowsPO-$\theta$ and MallowsPO-$\phi$ fine-tuned models are not only knowledgeable and but also able to deliver deeper and further insights, rather than focusing solely on the length. 


Finally, we also examine the models' performance in troubleshooting, code generation and debugging tasks. Tables \ref{tab:se_1_Mallows_theta_DPO_vs_BT_DPO} reveals that MallowsPO-$\theta$ fine-tuned model does better in identifying the potential problems in the JavaScript codes provided and providing more relevant, actionable suggestions. Additionally, Table \ref{tab:se_5_Mallows_phi_DPO_vs_BT_DPO} shows how MallowsPO-$\phi$ complements the suggestions by DPO by offering code snippets to further support the user.

\newpage
\section{Relevant Future Research}\label{app: future research}

For practical relevance, we consider the following directions on further leveraging these insights in real-world applications, focusing on curriculum learning and personalization through contextual scaling.

\textbf{Curriculum learning.}\hspace{2mm} Curriculum learning is a training strategy inspired by human learning processes, where models are trained on tasks or data organized from simpler to more complex examples, instead of presenting all examples randomly or simultaneously. This approach has been shown to guarantee faster training convergence \citep{bengio2009curriculum}, better generalization \citep{graves2017automated} and improved model robustness \citep{weinshall2018curriculum}. Within our framework, our proposed dispersion index can be utilized to indicate the determinism or diversity of the response to different questions, offering guidance on the complexity of the prompts to achieve curriculum learning. 

\textbf{Personalized Alignment.}\hspace{2mm} Furthermore, by considering $\phi(x, p)$, where $p$ is a personal feature, our method can be adjusted to consider a personalized preference ranking model. Note that how to achieve this with DPO is not straightforward. However, it is more natural in MallowsPO because of the existence of dispersion parameter $\phi(x)$ in the Mallows model, which controls how spread out the distribution should be. In this context, dispersion can be extended beyond the prompt level to encompass different users or user groups $p$. In terms of the preference optimization objective, different contextual scaling on the penalization coefficient can be considered to accommodate varying user or user group preferences for responses generated by the Supervised Fine-Tuning (SFT) model. Contextual scaling, tailored to different user preferences, enables a customized user experience. 

Both curriculum learning and personalization alignment \citep{chen2024pad} through contextual scaling present promising directions for further development and can be expanded based on our foundational research. However, these ideas merit dedicated presentation and discussion in separate works and are thus reserved for future exploration.

\clearpage

\end{document}